\PassOptionsToPackage{authoryear,round}{natbib}
\documentclass[unnumsec,webpdf,modern,large]{oup-authoring-template}%

\usepackage{upquote}
\usepackage{longtable}
\usepackage{bbm}
\usepackage{arydshln}
\graphicspath{{Fig/}}

\newcommand{\figref}[1]{Fig.~\ref{#1}}
\newcommand{\figpanelref}[2]{Fig.~\ref{#1}#2}
\newcommand{\tabref}[1]{Table~\ref{#1}}

\newcommand{\sifigref}[1]{SI Appendix, Fig.~S\ref{#1}}
\newcommand{\sitabref}[1]{SI Appendix, Table~S\ref{#1}}
\newcommand{\sisecref}[1]{SI Appendix, Section~\ref{#1}}

\theoremstyle{thmstyleone}%
\theoremstyle{thmstyletwo}%
\theoremstyle{thmstylethree}%

\newif\ifpreprint
\preprinttrue

\makeatletter
\newcommand{\ifpreprintproduction}[2]{\ifpreprint#2\else#1\fi}
\newcommand{\setpreprintpagestyle}{%
  \def\societylogo{}%
  \def\ps@opening{%
    \def\@oddhead{\raisebox{-10.3mm}[0pt][0pt]{\hbox to \textwidth{%
      \parbox{290pt}{\raggedright\sffamily\fontsize{7bp}{9bp}\selectfont
      \ifx\@appnotes\@empty\else\textbf{\@appnotes}\par\fi}\hfil}}}%
    \let\@evenhead\@oddhead
    \def\@oddfoot{\hfil\thepage\hfil}%
    \let\@evenfoot\@oddfoot}%
  \def\ps@headings{%
    \let\@oddhead\@empty
    \let\@evenhead\@empty
    \def\@oddfoot{\hfil\thepage\hfil}%
    \let\@evenfoot\@oddfoot}%
  \pagestyle{headings}%
}
\makeatother
\begin{document}

\ifpreprint\setpreprintpagestyle\fi

\ifpreprintproduction{\journaltitle{PNAS Nexus}}{\journaltitle{}}
\ifpreprintproduction{\DOI{DOI added during production}}{\DOI{}}
\ifpreprintproduction{\copyrightyear{YEAR}}{\copyrightyear{}}
\ifpreprintproduction{\pubyear{YEAR}}{\pubyear{}}
\ifpreprintproduction{\vol{XX}}{\vol{}}
\ifpreprintproduction{\issue{x}}{\issue{}}
\ifpreprintproduction{\access{Published: Date added during production}}{\access{}}
\ifpreprintproduction{\appnotes{Paper}}{\appnotes{Working Paper}}

\firstpage{1}


\title[Frontier AI performance across the business disciplines]{Frontier AI performance across the business disciplines: a case-grounded benchmark of knowledge work and analytical reasoning}
\makeatletter
\let\papertitle\@title
\makeatother

\author[1,$\ast$]{Ajay Patel\ORCID{0009-0003-1655-8703}}
\author[1]{Kartik Hosanagar\ORCID{0000-0002-6442-9434}}
\author[2]{Ramayya Krishnan\ORCID{0000-0001-9935-2468}}
\author[3]{Chris Callison-Burch\ORCID{0000-0001-8196-1943}}
\author[4]{Karim Lakhani\ORCID{0000-0002-5535-8304}}

\address[1]{\orgdiv{Operations, Information and Decisions Department}, \orgname{The Wharton School, University of Pennsylvania}, \orgaddress{\street{Philadelphia}, \postcode{19104}, \state{PA}, \country{USA}}}

\address[2]{\orgdiv{Heinz College of Information Systems and Public Policy}, \orgname{Carnegie Mellon University}, \orgaddress{\street{Pittsburgh}, \postcode{15213}, \state{PA}, \country{USA}}}

\address[3]{\orgdiv{Department of Computer and Information Science}, \orgname{University of Pennsylvania}, \orgaddress{\street{Philadelphia}, \postcode{19104}, \state{PA}, \country{USA}}}

\address[4]{\orgdiv{Technology and Operations Management Unit}, \orgname{Harvard Business School, Harvard University}, \orgaddress{\street{Boston}, \postcode{02163}, \state{MA}, \country{USA}}}

\corresp[$\ast$]{To whom correspondence should be addressed: \href{email:me@ajayp.app}{me@ajayp.app}}

\received{Date}{0}{Year}
\revised{Date}{0}{Year}
\accepted{Date}{0}{Year}


\abstract{Large language models (LLMs) are improving rapidly as reflected in benchmark scores, yet these AI benchmarks largely test capabilities such as factual recall, narrow question answering, mathematical problem-solving, and coding and agentic tool-use. What remains poorly measured is AI progress on the analytical knowledge work white-collar professionals perform daily, including synthesizing complex information, exercising judgment under uncertainty and incomplete information, applying strategic and adversarial thinking in multi-stakeholder settings, weighing trade-offs, and producing defensible, structured analyses. This gap is even more pronounced for subjective components of such work, where success can be challenging to define. The ``case method'' form of education practiced by top business schools provides a natural foundation for addressing this measurement gap, and we construct BusinessCaseBench, a benchmark spanning hundreds of questions drawn from business cases across eighteen disciplines, each paired with a grading rubric derived from the expert-written instructor case solution. On BusinessCaseBench, frontier AI models already score highly against instructor rubrics, and capability within one model family improves substantially over two years. These results provide strong evidence that AI performance on this class of work is already high and rapidly improving, with implications for business schools, where case pedagogy trains undergraduates and MBAs in this kind of analytical reasoning, and for entry-level professional roles, where such skills have historically anchored early-career work.} 

\keywords{knowledge work, analytical reasoning, business, education, benchmark, AI, large language models}

\otherabstract[Significance statement]{Business school trains students in decision-making and analytical reasoning: reading financial statements, evaluating market and competitive landscapes, designing operations, assessing strategy—shaping analysts, managers, and leaders in business and white-collar roles. We show frontier AI models already perform these skills at a high level across eighteen business disciplines, graded against expert-written standards. A longitudinal comparison within one model family documents roughly a 23-percentage-point gain over two years. We contribute BusinessCaseBench, a validated, discipline-spanning benchmark for AI in business reasoning and knowledge work that shifts the empirical question: not whether AI can do the kind of work business education trains for, but how business education and professional roles may change as analytic capability redistributes between humans and machines.}

\maketitle

\begin{figure*}[!t]
\centering
\includegraphics[width=\textwidth]{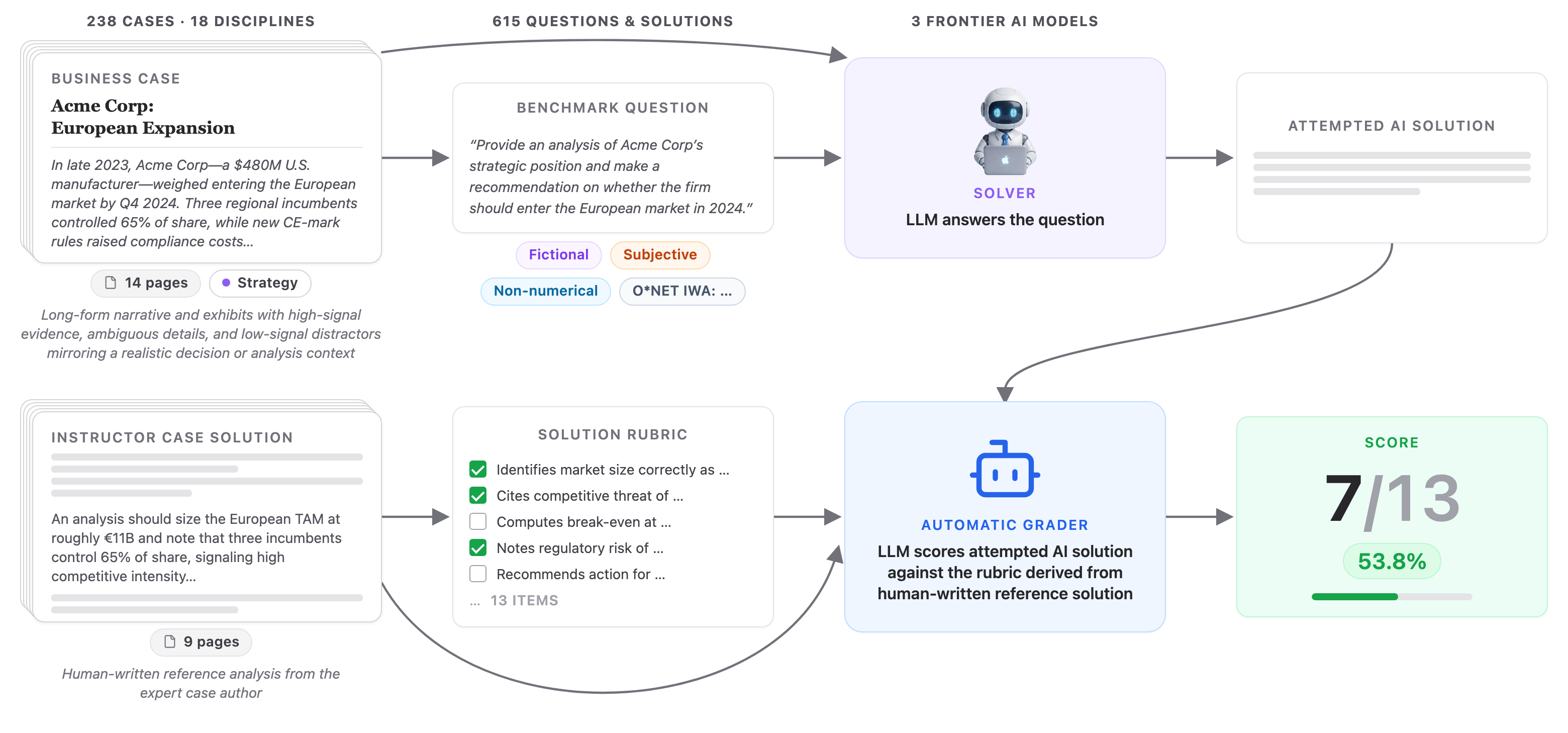}
\caption{The evaluation pipeline used to construct and score BusinessCaseBench. Case narratives and open-ended questions are paired with expert-written reference solutions from the instructor case solution. The reference solutions are transformed into equally-weighted checklist rubrics. A frontier AI model receives the case and question, produces an attempted solution, and an LLM-as-judge model scores the solution against each rubric criterion compared to the reference solution. Scores aggregate to Standard scoring (partial credit) and Complete Answer scoring (all criteria satisfied) metrics reported throughout the paper. We use human annotators to validate the automatic grading.}
\label{fig:pipeline}
\end{figure*}
\section{Introduction}\label{introduction}

Large language model (LLM) benchmark scores have risen rapidly \citep{gpt3_few_shot_learners,singh_openai_2026}, yet the benchmarks driving those scores largely assess a narrow set of well-defined competencies: factual recall, narrow question answering, mathematical problem-solving, and coding and agentic tool-use \citep{joshi_triviaqa_2017,zellers_hellaswag_2019,hendrycks_measuring_2021,chen_evaluating_2021,hendrycks_measuring_2021-1,jimenez_swe-bench_2024,merrill_terminal-bench_2026}. That design is well suited when correctness is unambiguous and answers can be checked against known facts, computed values, or enumerated choices. This pattern fails to capture, however, the capabilities required in the kind of analytical knowledge work white-collar professionals perform every day. That work calls for synthesizing complex information, exercising judgment under uncertainty and incomplete information, strategic and adversarial reasoning in multi-stakeholder settings, weighing trade-offs, and producing defensible, structured analyses---especially where success depends on subjective judgment and reasoning through multiple options rather than a single verifiable answer. This is the kind of work business schools train students to perform and is economically valuable to organizations and firms yet is scarcely measured in existing benchmarks \citep{patwardhan_gdpval_2025,wang_how_2026}.

A close analogue in medicine illustrates the same measurement gap and a constructive precedent. High LLM scores on the United States Medical Licensing Examination (USMLE) led many to infer that LLMs had matched clinical reasoning, yet USMLE questions are multiple-choice proxies optimized for administrative scale rather than for the open-ended synthesis high-performing clinicians perform on real cases. \citet{nori_sequential_2025} instead evaluated models on \textit{New England Journal of Medicine} case challenges---clinically demanding and expert-written diagnostic puzzle narratives with eventual solutions---using each case and its resolution to test reasoning under conditions closer to authentic practice with all of the ambiguity and uncertainty of real clinical work. Frontier models performed strongly on this design, even though such case challenges are meant to be rare and difficult to solve, leading to new insights about the strength of medical reasoning in frontier AI. The formulation validated expert-written professional cases as a more challenging and realistic alternative to narrow exam-style benchmarks like USMLE.

Business school case studies occupy a similar role in professional education that NEJM medical case challenges occupy in clinical training and education. They are curated, high-stakes, expert-written narratives that place learners in realistic decision contexts that require synthesis of complex information, judgment under uncertainty and incomplete information, strategic reasoning across multiple stakeholders, weighing trade-offs, and defensible structured analysis rather than narrow memorization or factual recall, as illustrated in \tabref{tab:representative-questions}. Publisher licensing of these cases and their use in the educational setting also limits pre-training exposure and contamination risk, making these cases a strong candidate for evaluation \citep{Xu2024BenchmarkDC}. These cases are often created by business school professors, with experience in conducting field studies, to describe a historical situation at a real organization or firm or a hypothetical situation at a fictional firm to train students for a particular business discipline \citep{rebeiz_insider_2011,Nohria2021CaseMethod}. Prior efforts to benchmark LLMs on business-relevant tasks have remained narrowly scoped, with some targeting numerical reasoning within finance \citep{chen_finqa_2022,koncel-kedziorski_bizbench_2024,wu_bloomberggpt_2023,xie_finben_2024}, others evaluating domain-adjacent language tasks such as format-following or ad-copy generation \citep{xia_fofo_2024,liu_llms_2025,wang_enterprise_2025}, and a growing line of work assessing LLM agents operating in enterprise software systems or domain-specific interactive environments \citep{drouin_workarena_2024,boisvert_workarena_2025,huang_crmarena_2025,huang_crmarena-pro_2025,li_investorbench_2024,xu_theagentcompany_2025,patwardhan_gdpval_2025}. A few recent works have measured AI performance with business cases or simulations \citep{ai_biz_impact_3, dellacqua2026jagged,allen2026strategy}, but are limited in their breadth of coverage across business disciplines and kinds of knowledge work. These prior benchmarks and studies provide useful signal on specific capabilities relevant to business and knowledge work, but none measures open-ended analytical work across the full breadth of business disciplines under conditions that mirror authentic professional judgment.

We extend case-grounded evaluation to business disciplines through a design pattern and automated pipeline for benchmarking complex analytical knowledge work. Using business school case studies paired with reference solutions derived from instructor case solutions, we generate structured, equally-weighted checklist rubrics. Model responses are then evaluated against those rubric criteria for scoring, following the pipeline illustrated in \figref{fig:pipeline}. These evaluations yield BusinessCaseBench, a discipline-spanning capability map of frontier LLMs across 615 questions drawn from 238 cases spanning eighteen business disciplines. Each question is additionally mapped to the O*NET occupational taxonomy \citep{onet_online_2026}, so results can be read by professional work activity and occupation and used to characterize where implied AI impact concentrates. Finally, a four-model within-family generational trajectory documents how performance has evolved over two years.

Several findings follow for how business education and labor markets should adapt as frontier AI capability on this class of work rises. Top frontier models achieve rubric-graded accuracy above 87\% under partial credit scoring, suggesting that aggregate performance on these tasks, many of which involve reasoning and judgment over uncertain and ambiguous information, is now high. Improvement across model generations is broad-based, with approximately a 23-percentage-point gain within one model family over two years, and is not confined to quantitative subtasks where older models previously had the most room to improve \citep{hendrycks_measuring_2021-1}. Discipline is associated with far wider differences in model difficulty than question type, with gaps between questions from fictional and real cases, numerical and non-numerical questions, and subjective and objective questions each small relative to the variation across disciplines. Finally, genuine capability ceilings are rare. Fewer than 7\% of questions defeat every frontier model, indicating that the required knowledge is distributed across the frontier rather than absent from it and that complete and thorough responses to highly open-ended analytical tasks remain the main challenge and opportunity for human-AI collaboration.


\begin{table*}[!t]
\centering
\caption{Representative examples from BusinessCaseBench of case questions.}\label{tab:representative-questions}%
\small
\begin{tabular*}{\textwidth}{@{\extracolsep\fill}>{\raggedright\arraybackslash}p{0.13\textwidth}>{\raggedright\arraybackslash}p{0.31\textwidth}>{\raggedright\arraybackslash}p{0.13\textwidth}>{\raggedright\arraybackslash}p{0.31\textwidth}@{\extracolsep\fill}}
\hline
Case Discipline & Representative Question Summary & Case Discipline & Representative Question Summary \\
\hline
Business \& Government Relations &
Write a Supreme Court judicial opinion that rules on a patent settlement between a branded pharmaceutical firm and a generic manufacturer, and establishes a legal framework lower courts can apply to similar agreements. &
Leadership \& Organizational Behavior &
Advise a hospital's chief medical officer on whether to promote a division chief whose clinical excellence conflicts with the institution's collaborative culture. \\
Economics &
As a municipal consultant, analyze participatory-budgeting voting data from a prior city program to evaluate whether approval voting produces fair, proportional outcomes and recommend a voting rule for a new initiative. &
International Business &
Advise an executive at a multinational's overseas subsidiary navigating conflicting demands from headquarters and local workplace norms, proposing a strategy that delivers measurable results while respecting local cultural practices. \\
Business Ethics &
Analyze a data scientist's ethical stance when asked to add a feature that estimates pregnancy likelihood for female job applicants to AI hiring software. &
Finance &
Build a discounted cash flow valuation of a luxury automaker at its public offering and quantify how equity value changes under a sharp currency depreciation and alternative management responses. \\
\hline
\end{tabular*}
\end{table*}

\section{Overview}\label{overview}

BusinessCaseBench evaluates open-ended analytical knowledge work using 238 business school cases paired with instructor solutions, yielding 615 questions across eighteen disciplines from ``Strategy'' and ``Finance'' to ``Leadership \& Organizational Behavior,'' with the full compositional breakdown reported in \sisecref{supp:composition-tables}. The benchmark spans subjective and open-ended analytical questions as well as objective, numerical reasoning questions. We classify each question as derived from a case about a fictional firm or a real firm, as well as whether the question is numerical or non-numerical, and whether it is subjective or objective. The benchmark includes 245 questions on fictional firms and 370 on real firms, 184 numerical and 431 non-numerical questions, and 339 subjective and 276 objective questions. Each question is mapped to the O*NET occupational taxonomy developed by the U.S. Department of Labor \citep{onet_online_2026} at the Work Activity (WA), Intermediate Work Activity (IWA), and Detailed Work Activity (DWA) levels. Together these labels cover 24 WAs, 55 IWAs, and 108 DWAs, so results can be read against professional work classifications as well as academic disciplines; see \sisecref{supp:composition-tables} for tabular summaries of these mappings.

For every question, the instructor case solutions supply a reference solution that we extract and use to generate a checklist-style rubric, as illustrated in \figref{fig:pipeline}. A model receives the full case narrative and exam-style question prompt and produces an open-ended attempted solution. The attempted solution is then evaluated with an LLM-as-judge protocol \citep{zheng_judging_2023} against rubric criteria using the reference solution. We report two complementary scores throughout. Standard scoring awards partial credit as the rubric-weighted fraction of criteria satisfied. It captures how much of what the instructor case solution expected the model actually produced, even when not every element is present. Complete Answer scoring, shown in \figref{fig:complete_answer}, is stricter and all-or-nothing. It marks whether a model satisfies every criterion on a question's rubric. Standard scoring summarizes how much of the instructor rubric a response covers and, especially on open-ended and subjective questions, often provides the more faithful gauge of answer quality when full checklist satisfaction is neither required nor always attainable. Complete Answer scoring complements it by recording how often responses nonetheless satisfy every criterion---a conservative completeness check when partial credit scores are already high. Precise formal definitions and the scoring protocol appear in the Methods section and \sisecref{supp:model-evaluation-protocol}. For a subset of questions, three trained annotators independently authored a rubric and graded each assigned response before seeing any automated rubric or score. This blinded protocol checks whether the evaluation, automated rubrics, and automated grading accurately measure an attempted solution's success or failure in a way that directionally correlates with human judgement; see the Methods section and \sisecref{supp:human-annotation-validation} for further details.

Primary comparisons use three frontier AI models, OpenAI GPT-5.4 \citep{singh_openai_2026}, Anthropic Claude Sonnet~4.6 \citep{anthropic_system_2025}, and Google Gemini~3 Flash Preview \citep{deepmind_gemini_2025}, evaluated on all 615 questions with pinned model identifiers and sampling settings documented in \sisecref{supp:model-evaluation-protocol}. A within-family generational analysis evaluates four successive OpenAI models spanning roughly two years (GPT-4 Turbo to GPT-5.4) on the same fixed benchmark \citep{openai_gpt-4_2024}. This comparison tracks how AI performance on this class of work has evolved over roughly two years of model development.

\begin{figure*}[!t]
\centering
\includegraphics[width=\textwidth]{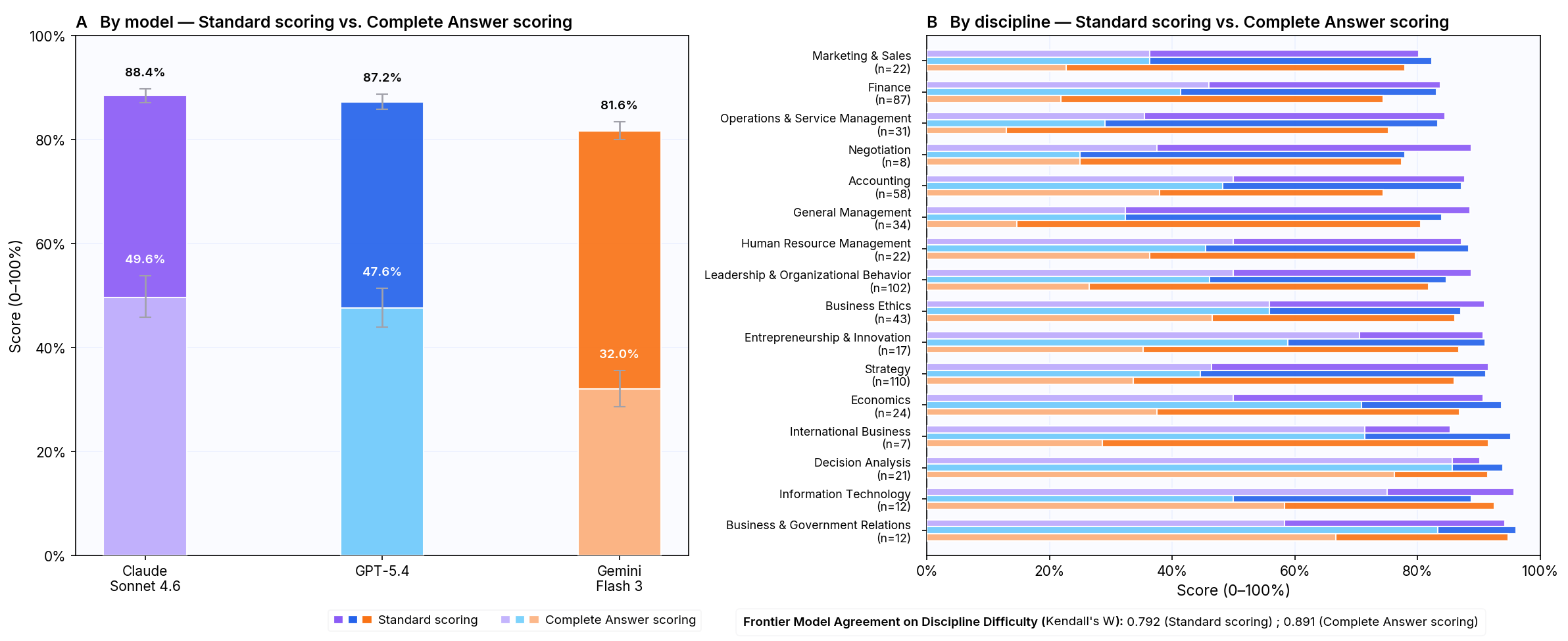}
\caption{Frontier model performance under Standard scoring and Complete Answer scoring. \textbf{A)} Mean scores for GPT-5.4, Claude Sonnet~4.6, and Gemini~3 Flash Preview on all 615 BusinessCaseBench questions; bars show Standard scoring (full height) with Scores under Complete Answer scoring overlaid (lighter segment). Error bars are bootstrapped 95\% confidence intervals. \textbf{B)} The same two metrics by business discipline (only showing disciplines with $n \geq 5$ questions); disciplines are ordered by ascending mean Standard scoring. Scores under Complete Answer scoring are uniformly lower than those under Standard scoring and widen the spread across models and disciplines, indicating that high partial credit often coexists with incomplete multi-criterion satisfaction.}
\label{fig:complete_answer}
\end{figure*}
\section{AI performance across the full business canon}\label{ai-performance-across-the-full-business-canon}

Across the full 615-question benchmark, the organizing pattern is broad competence under partial credit, much lower full rubric satisfaction at the same scale, and a discipline ordering that all three models share. \figref{fig:complete_answer} summarizes this structure for GPT-5.4, Claude Sonnet~4.6, and Gemini~3 Flash Preview under Standard and Complete Answer scoring.

\subsection{Aggregate performance}\label{aggregate-performance}

Under Standard scoring, all three frontier models perform at a uniformly high level on the full benchmark. \figpanelref{fig:complete_answer}{A} places Claude Sonnet~4.6 at 88.4\% (95\% CI [87.1, 89.7]), GPT-5.4 at 87.2\% (95\% CI [85.8, 88.7]), and Gemini~3 Flash Preview at 81.6\% (95\% CI [80.0, 83.4]). The top-to-bottom spread is only 6.8 percentage points, and confidence intervals for the two leading models overlap throughout. These scores indicate that, when rubric criteria are credited independently, frontier models already cover most of what instructor case solutions expect on open-ended business case work spanning the full discipline canon.

\subsection{AI outputs as drafts, not verdicts}\label{partial-versus-complete-answers}

High partial credit scores hide a stricter picture. Complete Answer scoring requires every rubric criterion on a question to be satisfied, and full satisfaction is far rarer. In \figpanelref{fig:complete_answer}{A}, the lighter overlaid bars show that Claude Sonnet~4.6 reaches it on 49.6\% of questions (95\% CI [45.9, 53.8]), GPT-5.4 on 47.6\% (95\% CI [43.9, 51.4]), and Gemini~3 Flash Preview on 32.0\% (95\% CI [28.6, 35.6]). Even the leading model leaves more than half of questions without a fully complete answer by instructor standards. The top-to-bottom spread widens to 17.6 percentage points, roughly 2.6 times the spread under Standard scoring, because a response can earn high partial credit while still missing one or more required elements. Complete Answer scoring is intentionally a conservative lower-bound measure of performance in this sense. Any missed criterion fails the question, even when the satisfied portion of the rubric would already support a strong partial credit grade that many instructors might treat as analytically sufficient on subjective, open-ended questions where full rubric satisfaction might be an unreasonably strict expectation. High Standard scores therefore describe analytically strong drafts that capture much of the expected analysis, not outputs that meet the full scope of the instructor rubric without review.

\subsection{Discipline-level variation}\label{discipline-level-variation}

Discipline-level performance is structured rather than noisy. \figpanelref{fig:complete_answer}{B} orders the sixteen disciplines with at least five questions by ascending mean Standard score. \sitabref{tab:discipline-model-matrix} reports the underlying cross-model means: Standard scores range from 80.1\% in Marketing \& Sales to 95.0\% in Business \& Government Relations, a spread that exceeds the gap among the three models themselves, and under Complete Answer scoring the same disciplines separate more sharply, from 25.8\% in Operations \& Service Management to 82.5\% in Decision Analysis. All three frontier models rank disciplines in nearly the same order, with high rank concordance (Kendall's $W$ = 0.79 under Standard scoring and 0.89 under Complete Answer scoring \citep{kendall_problem_1939}), indicating that difficulty reflects discipline-specific task structure more than any single provider's training idiosyncrasies.

\begin{figure*}[!t]
\centering
\includegraphics[width=\textwidth]{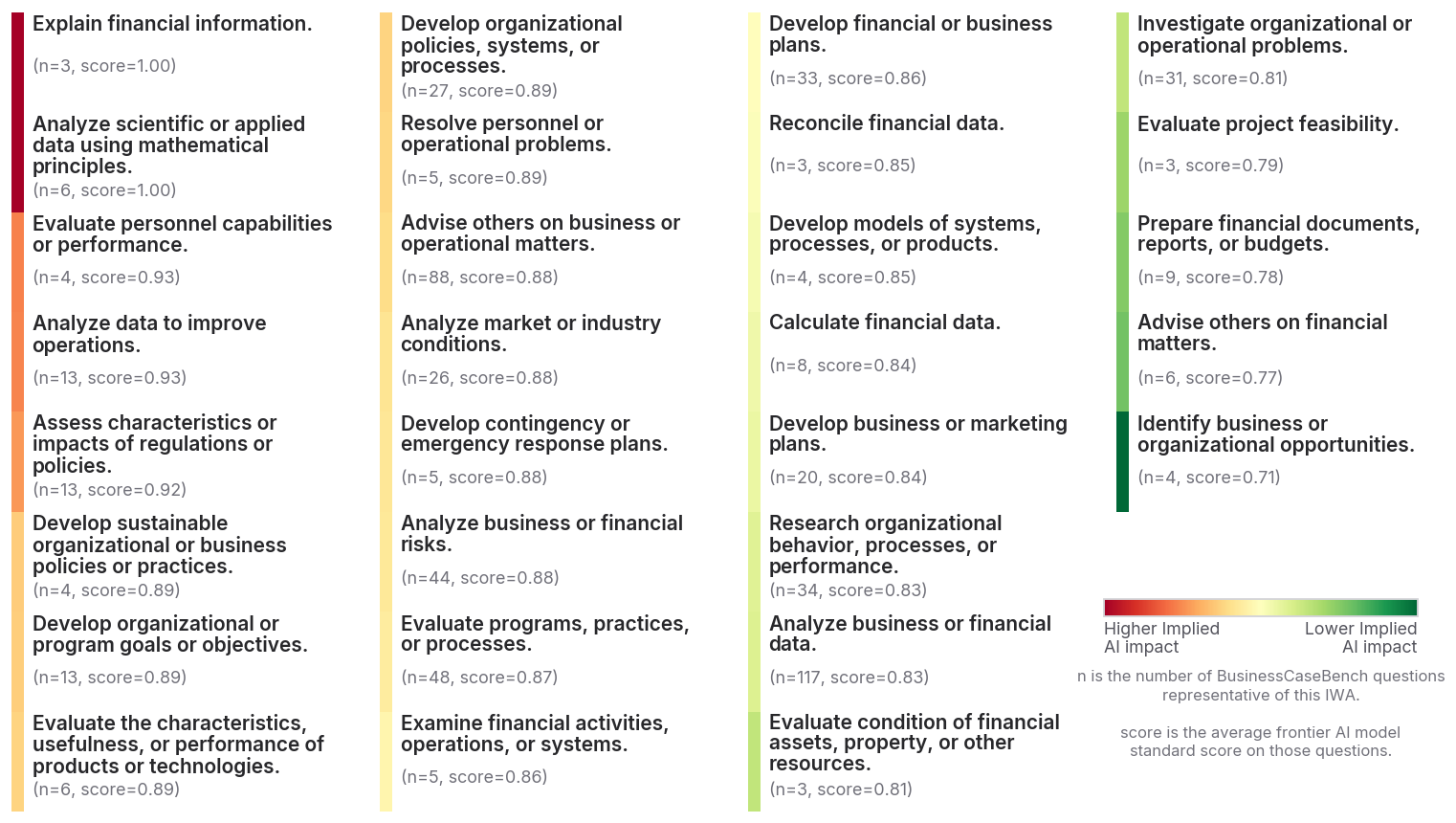}
\caption{Frontier AI performance by O*NET Intermediate Work Activities (IWAs). Each cell is an IWA with at least three BusinessCaseBench questions ($n \geq 3$); color encodes the mean score under Standard scoring averaged across GPT-5.4, Claude Sonnet~4.6, and Gemini~3 Flash Preview on those questions (red tones indicate higher implied AI impact to those work activities). IWAs are sorted by implied AI impact. Open-ended advisory and opportunity-identification activities (e.g., identifying organizational opportunities, advising on financial matters) rank among the hardest; structured analytical and explanatory activities approach ceiling performance.}
\label{fig:iwa_heatmap}
\end{figure*}

\section{Characterizing model capabilities and strengths}\label{characterizing-model-capabilities-and-strengths}

Aggregate scores hide how case framing, question metadata, and cross-model patterns organize the remaining gap. We therefore characterize what predicts residual difficulty and how benchmark questions map onto professional work activities.

\subsection{Fictional cases are slightly easier than real ones}\label{fictional-cases-are-slightly-easier-than-real-ones}

Questions from real cases score slightly higher than questions from fictional cases on the full benchmark (86.5\% vs.\ 84.6\% under Standard scoring, a 1.9 percentage-point gap), a pattern attributable largely to a discipline-specific outlier in Finance rather than a uniform fiction--real effect. In six of the nine disciplines with sufficient questions in both arms, questions from fictional cases score at or above questions from real cases, consistent with real narratives introducing longer contexts, more ambiguity, and extraneous detail that complicate synthesis and make for a more challenging case question. Finance and Accounting reverse the pattern: fictional finance questions score 15.0 percentage points below real finance questions (70.9\% vs.\ 85.9\%), and Accounting shows a smaller 2.7-point gap in the same direction, with Finance standing out as the lone large negative outlier in \sifigref{fig:fiction_vs_real_delta}. We hypothesize that questions from real cases in these disciplines often anchor on named companies whose financial profiles may appear in public filings or financial press, giving models a pre-training edge that questions from fictional cases, which strip away that exposure, do not provide. The Finance outlier therefore likely reflects how pre-training interacts with case framing rather than a general advantage of fictional material.

\subsection{Difficulty is largely case- and question-specific}\label{difficulty-is-largely-case-and-question-specific}

Characterizing what makes a question hard largely resists coarse categorical metadata like question type or discipline, and most of the variation in difficulty lies beyond these labels in demands specific to a given case or question. \sifigref{fig:question_strata} places mean performance across the six question-type strata in a narrow 82.3--87.2\% band under Standard scoring. Numerical questions score lowest (82.3\%) and non-numerical highest (87.2\%), subjective and objective questions differ by only 0.8 percentage points, and questions from fictional versus real cases differ by 1.9 points when pooled across all questions, although we note earlier that for some disciplines the gap is larger. Complete Answer scoring widens these gaps by question type modestly but leaves them small relative to the spread across disciplines that \figref{fig:complete_answer} shows. A regression of mean frontier Standard score on question-type strata and discipline categories, using effect coding with disciplines of fewer than five questions pooled into a single Other category and with standard errors clustered by case, confirms that these coarse labels explain little of the variation in scores. Question-type tags alone account for 2.6\% of question-level score variation (adjusted $R^{2}$), discipline alone for 3.6\%, and both together for 5.1\%, and under cross-validation the two blocks predict held-out cases comparably. Each block still carries a detectable association with difficulty. Adding discipline after question type is jointly significant ($p < 0.001$), as is adding question type after discipline ($p = 0.007$). Once discipline and the other strata are accounted for, scores on fictional and real cases no longer differ in a meaningful way, and numerical questions remain modestly harder ($-$5.9 points). Most of the difference between harder and easier questions therefore lies in case- and question-specific demands that coarse question-type and discipline metadata do not capture, such as longer ambiguous narratives, multi-part rubric structure, and more open-ended advisory or evaluative reasoning. A variance decomposition across the $238$ source cases supports that reading. Questions from the same case have moderately correlated difficulty (intraclass correlation $0.22$), knowing which case a question comes from predicts its score far better than the coarse metadata labels do (adjusted $R^{2}=0.22$ vs.\ $0.05$), and nearly half of the variation in scores occurs among questions within the same case. Full nested-model statistics, adjusted $R^{2}$, cross-validated comparisons, and the case-level decomposition appear in \sisecref{supp:metadata-regression}.

\subsection{Difficulty is rarely absolute}\label{difficulty-is-rarely-absolute}

Few benchmark questions defeat every evaluated model. \sitabref{tab:oracle-hardness} reports that only 43 of 615 questions (7.0\%) have a maximum Standard score at or below 70\% across GPT-5.4, Claude Sonnet~4.6, and Gemini~3 Flash Preview, and that only 521 of 8,369 rubric criterion instances (6.2\%) never receive credit from any of the three. A cross-model oracle that takes, per question, the best score among the three frontier models reaches 92.8\%, 4.5 percentage points above the leading single model. What one model misses on a given question is therefore often captured by another, so remaining error reflects incomplete satisfaction of multi-part rubrics more often than absolute capability ceilings; \sisecref{supp:cross-model-complementarity} documents the underlying complementarity pattern.

\subsection{From cases to an occupational map}\label{from-cases-to-an-occupational-map}

\figref{fig:iwa_heatmap} maps benchmark questions to O*NET Intermediate Work Activities, bridging academic disciplines to occupational work activities and occupations to assess the implied AI impact that BusinessCaseBench would predict. Across 29 Intermediate Work Activities (IWAs) with at least three benchmark questions, mean Standard scores range from 70.9\% on ``Identify business or organizational opportunities'' to ceiling performance on bounded explanatory and mathematical analysis activities. The hardest activities combine open-ended advisory framing with case-specific quantitative or evaluative reasoning, whereas well-scoped data interpretation and performance evaluation tasks approach saturation. Because O*NET's work activity taxonomy involves overlapping and partially redundant categories at different levels of granularity \citep{onet_limitations}, we treat these mappings as indicative rather than definitive. Implied AI impact on occupations, which \sifigref{fig:ai_impact_on_onet_occupations} provides, should be read alongside the coverage bands in that figure that show how thoroughly each job's relevant work activities are exercised in the benchmark.

\begin{figure*}[!t]
\centering
\includegraphics[width=\textwidth]{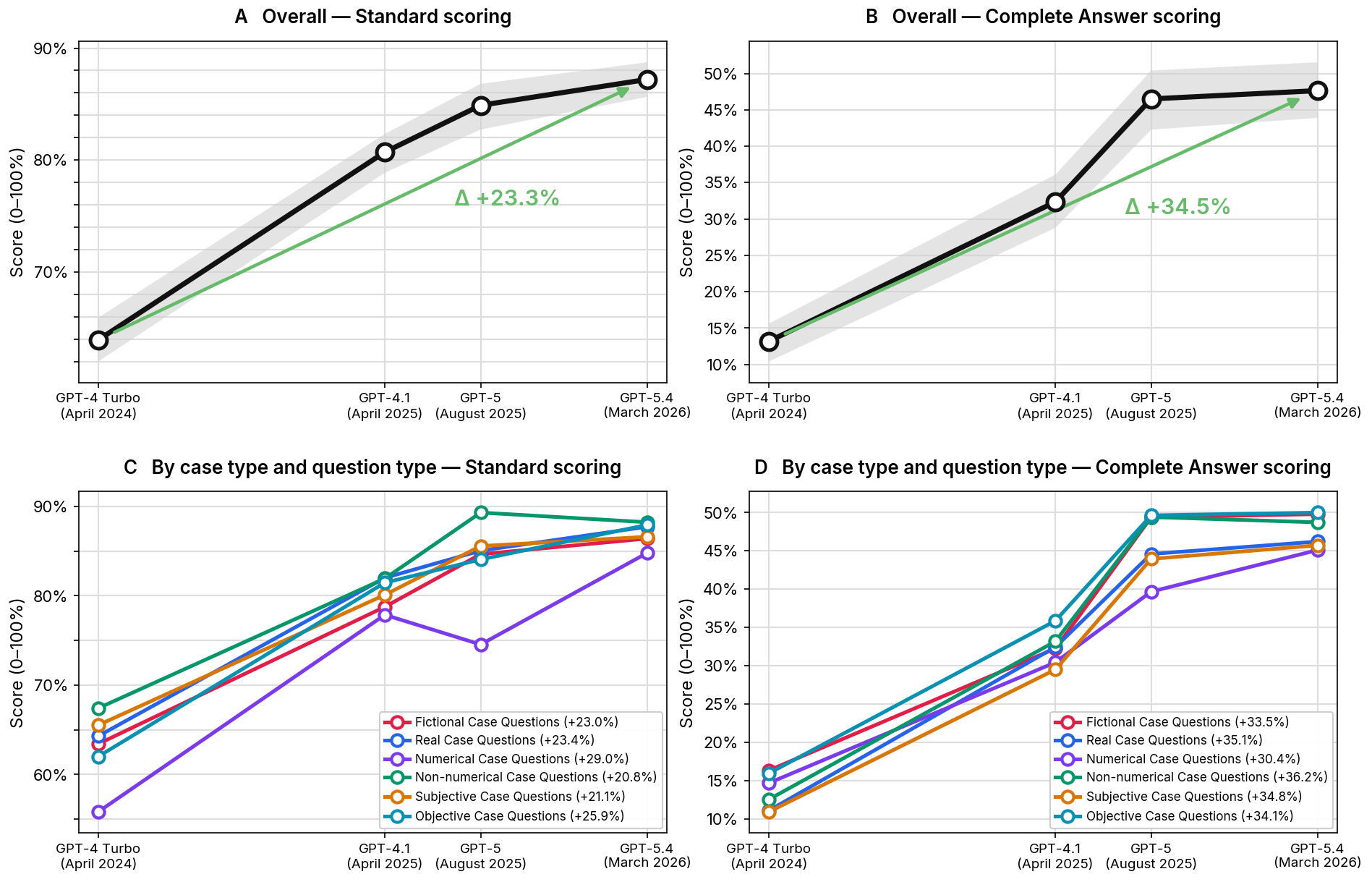}
\caption{Within-family performance trajectory for four OpenAI models evaluated on the fixed 615-question BusinessCaseBench (GPT-4 Turbo, GPT-4.1, GPT-5, and GPT-5.4; $\approx$ two years). \textbf{A)}  Standard scoring rubric-weighted scores with 95\% bootstrap confidence bands; aggregate gain from earliest to latest model is annotated. \textbf{B)} Scores under Complete Answer scoring on the same models. \textbf{C)} Standard scoring decomposed by case type and question type (fictional vs.\ real cases; numerical vs.\ non-numerical; subjective vs.\ objective). \textbf{D)} Scores under Complete Answer scoring for the same six strata. All strata improve from GPT-4 Turbo to GPT-5.4.}
\label{fig:openai_trajectory}
\end{figure*}
\section{A generational comparison within one model family}\label{a-generational-comparison-within-one-model-family}

How quickly has AI capability on this class of work improved? Holding the benchmark fixed, we evaluate four successive OpenAI releases spanning roughly two years, from GPT-4 Turbo through GPT-4.1, GPT-5, and GPT-5.4, to trace the generational trajectory of performance on open-ended analytical knowledge work. We choose the OpenAI model family as we are able to reliably access all four models for retrospective evaluation.

\paragraph{Four-generation performance trajectory}\label{four-generation-performance-trajectory}

\figpanelref{fig:openai_trajectory}{A} and \figpanelref{fig:openai_trajectory}{B} summarize the trajectory on all 615 questions under Standard scoring and Complete Answer scoring. Mean Standard score rises from 63.9\% for GPT-4 Turbo (95\% CI [62.0, 65.9]) to 87.2\% for GPT-5.4 (95\% CI [85.8, 88.7]), a gain of 23.3 percentage points across the four models. The stricter Complete Answer score moves in parallel, from 13.2\% (95\% CI [10.7, 15.8]) to 47.6\% (95\% CI [43.9, 51.4]), a 34.4-point gain that shows improvement on full and complete answers, not only on partial credit. Intermediate releases step up the curve in sequence, with GPT-4.1 reaching 80.7\% under Standard scoring and GPT-5 crossing 84.9\% before the latest model approaches the high partial credit levels reported in \figref{fig:complete_answer}.

\paragraph{Categorizing the distribution of improvement}\label{categorizing-the-distribution-of-improvement}

The gains are broad-based rather than confined to easy slices of the benchmark. \figpanelref{fig:openai_trajectory}{C} and \figpanelref{fig:openai_trajectory}{D} decompose the same four-model trajectory by fictional versus real cases and by numerical, non-numerical, subjective, and objective questions. Under Standard scoring, fictional and real cases improve by essentially the same margin (23.0 and 23.4 percentage points), numerical questions by 29.0 points, and non-numerical questions by 20.8 points. Under Complete Answer scoring every stratum gains at least 30 points, and non-numerical and subjective questions improve as much as or more than their numerical and objective counterparts (for example, +36.2 points on non-numerical questions versus +30.4 on numerical ones). That pattern runs counter to a hypothesis in which progress is concentrated on quantitative reasoning alone, the kind of capability early models struggled with and that popular benchmarks measure---and that emphasis has led to improvements in numerical precision and mathematical reasoning \citep{hendrycks_measuring_2021-1}. \sifigref{fig:openai_trajectory_discipline} extends the same pattern to disciplines in panels A and B: all twelve disciplines with at least fifteen questions post double-digit Standard gains, including large moves in Economics (+31.7 points), Accounting (+28.8 points), and Finance (+28.7 points), where GPT-4 Turbo began well below today's levels. Even disciplines that were already relatively strong in 2024, such as Business Ethics and Strategy, continue to gain under Complete Answer scoring. The most recent step from GPT-5 to GPT-5.4 is smaller (+2.3 points on Standard scoring and +1.1 points on Complete Answer scoring), indicating slowing improvement, which is expected as scores approach the ceiling. In summary, the two-year arc documents a substantial gain in capability on the analytical work this benchmark measures.

\section{Extended results on the latest frontier of ultra-large models}\label{extended-results-latest-frontier-ultra-large-models}

After the primary experiments reported above were completed, Anthropic released Claude Fable~5, a Mythos-class model substantially larger than previously released public systems and marketed with particular emphasis on software engineering capability. Deployment was temporarily interrupted under U.S.\ government export controls related to cybersecurity concerns \citep{anthropic_redeploying_fable_2026}. As this release fell near the end of our evaluation period for this research, we report Claude Fable~5 as an after-the-fact Extended Result on BusinessCaseBench. The purpose of this extension is to document how a contemporaneous ultra-large frontier model performs on the open-ended knowledge-work tasks measured here.

Aggregate performance for Claude Fable~5 is nearly indistinguishable from Claude Sonnet~4.6. Under Standard scoring, Fable~5 attains 88.0\% (95\% CI [86.4, 89.5]), compared with 88.4\% for Sonnet~4.6 (\sitabref{tab:standard-leaderboard}). Under Complete Answer scoring, Fable~5 attains 50.9\% (95\% CI [47.3, 54.8]), a 1.3-percentage-point difference relative to Sonnet~4.6 at 49.6\% (\sitabref{tab:complete-leaderboard}). Confidence intervals for the two models overlap on both metrics. Such similarity is consistent with substantial overlap in training data between successive models within a provider family. \sitabref{tab:discipline-model-matrix} displays the discipline-level breakdown for Fable~5; across disciplines, Fable~5 remains close to Sonnet~4.6 and does not rearrange the harder and easier domains previously identified.

On BusinessCaseBench, Claude Fable~5's similar performance profile to Claude Sonnet~4.6 may reflect this release class's greater emphasis on agentic software engineering capabilities like cybersecurity. The Extended Result therefore leaves the study's qualitative conclusions unchanged. Frontier models already achieve high Standard scores, Complete Answer satisfaction remains substantially harder, and the latest ultra-large Anthropic release does not alter the capability picture this benchmark establishes for business knowledge work. One provisional conclusion that may be drawn, however, is that optimizing for software engineering workflows does not necessarily improve the capabilities of these models on this class of knowledge work.

\section{Discussion}\label{discussion}

Frontier AI models already perform well on open-ended business case questions graded against expert-written instructor case solutions, yet this economically central work remains weakly represented in the benchmarks that track LLM progress. On this benchmark, aggregate performance under Standard scoring is high, and capability gains within one model family over roughly two years are large, as the generational trajectory in \figref{fig:openai_trajectory} shows. Residual difficulty is characterized more by gaps on certain business disciplines and occupational activity than by simple categorization (numerical vs. non-numerical or subjective vs. objective questions). Few questions systematically defeat \textit{every} evaluated frontier AI model, and a cross-model oracle performs well above the best single model. These patterns suggest that the remaining gap reflects incomplete satisfaction of multi-part rubrics more often than truly missing domain knowledge, a distinction \figref{fig:complete_answer} makes especially visible under Complete Answer scoring, since even leading systems satisfy every rubric criterion on fewer than half of questions. Gaps in attempted solutions are frequently model-specific rather than absolute, and much of what one model misses may be captured by another model on the same question. The empirical question has shifted from whether frontier models can perform the analytic knowledge work business cases demand to how completely they do so, where they remain weakest, and how business schools and firms should respond as capability improves.

Our evaluation rests on the same curated artifacts professional schools use to train judgment, reasoning, and analytical skills. Treating expert-written cases as evaluation instruments is increasingly attractive across specialized domains because they approximate the real task more closely than convenient proxies like standardized exams do, because they carry realistic context, simulate information gaps and ambiguity, require judgment under uncertainty, and their long-form narratives and analytical solutions align more closely with realistic deployed LLM use than short multiple-choice/free-response questions found in standardized exams. \citet{nori_sequential_2025} demonstrated a challenging medical diagnosis and reasoning benchmark by using complex NEJM medical cases as a more realistic alternative than measuring medical reasoning with standardized exams like USMLE. Our study and benchmark extend that design pattern to business school cases to test analytical reasoning skills required in realistic knowledge work. In both works, license-restricted cases impose an openness cost, but they also strengthen validity because verbatim pretraining exposure is less plausible and high scores are harder to dismiss as memorization \citep{Xu2024BenchmarkDC}. In this evaluation regime, models earn credit for satisfying articulated expectations rather than matching a single, short gold string, which better mirrors a quality response, at the price of requiring rubrics that are well specified and stably scored.

The case format also lowers barriers to benchmark construction for specialized domains. Domain experts without technical training can describe authentic situations in the narrative form---the context at hand, what was known and unknown, the options weighed, trade-offs, and the reasoning behind a decision, captured in a reference solution. Once cases and solutions are written in that structure, automated extraction, rubric synthesis, and checklist grading can scale capability measurement across models and time, as \citet{nori_sequential_2025} and BusinessCaseBench both illustrate. More challenging benchmarks, in turn, help push model development further. This approach still requires validation that the automated rubrics and grading match human judgements of quality and skill---a step we take in our study through human annotation.

Several limitations bound these results and point to further evaluation work in this domain. While we do not find any evidence of verbatim contamination of case materials in common large pre-training corpora, frontier models are trained on proprietary datasets that may or may not have exposure to some case materials that would be difficult to verify without open access. As we discuss earlier, some cases based on real firms may hinder unbiased evaluation as similar data may be published under public market regulatory filings or in the financial press, although we find the effect to largely systematically affect a single discipline (Finance). As is commonly known, LLM-as-judge scoring retains residual risks, brittleness on long answers, and numerical lapses, even after human validation \citep{zheng_judging_2023}. In our grading, high partial credit under Standard scoring can coexist with failed completeness under Complete Answer scoring for cases where full-credit is unreasonably strict and a strong partial credit answer may be a sufficient baseline of expert human performance. Rubric-guided scoring generally assesses only whether each stated criterion is accurately satisfied, but it does not audit the response for extraneous, unsubstantiated, or factually incorrect content outside the checklist. The benchmark is in English, and other languages, institutional settings, and pedagogical norms may yield different profiles of skills and capability. The occupational mapping in \figref{fig:iwa_heatmap} is informative but incomplete both because O*NET work activities are broad and may only be loosely connected to some of the skills tested in a case and because many work activities appear rarely in the case corpus. Results should therefore be read alongside the occupation coverage bands shown in \sifigref{fig:ai_impact_on_onet_occupations}. The evaluation is also single-turn. Iterative clarification, information gathering, active negotiation, and formal accountability in firms lie outside what we measure and such areas may be opportunities for human-AI teaming \citep{ai_biz_impact_1}.

Progress on AI for knowledge work will require a portfolio of benchmarks that separate discrete skills from integrative judgment and tie models to economically meaningful tasks \citep{patwardhan_gdpval_2025,wang_how_2026}. We contribute BusinessCaseBench, a discipline-spanning case benchmark with rubric-graded scoring that sits between narrow quantitative/analytical suites \citep{chen_finqa_2022,koncel-kedziorski_bizbench_2024,xie_finben_2024} and agent benchmarks focused on enterprise software use \citep{drouin_workarena_2024,xu_theagentcompany_2025}. Performance against instructor case solution standards is already high and rising within at least one model family. As \figref{fig:iwa_heatmap} summarizes, models are comparatively strong on structured analytic tasks and more exposed on open-ended advisory and analysis tasks. They also often fall short of full rubric satisfaction under Complete Answer scoring even when partial credit under Standard scoring is substantial, a gap evident in \figref{fig:complete_answer}.

These findings from BusinessCaseBench have immediate implications for business education and white-collar labor. Case pedagogy trains the judgment, synthesis, and defensible reasoning that MBA programs and undergraduate curricula were designed to build, skills that have historically anchored early-career analytical work. With frontier models already scoring high under Standard scoring and improving quickly within at least one model family, producing a strong draft answer is increasingly inexpensive. The design problem for business schools is therefore less whether students can generate plausible case analyses and more how curricula cultivate verification, thoroughness, and the ability to recognize what a fully complete answer requires, especially on tasks where Standard scores are already high but Complete Answer satisfaction remains elusive \citep{ai_biz_impact_2}. For labor markets, the occupational mapping in \figref{fig:iwa_heatmap} and the occupation coverage bands in \sifigref{fig:ai_impact_on_onet_occupations} help identify where entry-level analytical work appears most exposed to AI. They also suggest where human advantage may remain, particularly in work that depends on integrating stakeholder perspectives, exercising accountability, and navigating organizational context. Such capabilities lie beyond the scope of single-turn benchmarks and are therefore undermeasured in evaluations such as ours.. Those remaining gaps are natural sites for human--AI teaming rather than wholesale substitution \citep{dellacqua2026jagged,krakowski2025human}. The within-family trajectory we document is a lower bound on how quickly capability can improve. It does not forecast every model provider's path, but it argues against treating today's Standard scores as a ceiling on what organizations and firms will soon see in practice.

BusinessCaseBench also has implications for AI evaluation and model development. As many widely used AI benchmarks approach saturation, measuring further progress will increasingly require evaluations grounded in economically meaningful knowledge work rather than narrow factual recall, mathematical reasoning, or coding tasks alone. BusinessCaseBench identifies a subset of open-ended analytical reasoning that remains challenging for frontier models despite their strong overall performance. Models often produce analytically strong drafts while omitting important considerations, failing to integrate competing constraints, or stopping short of a complete recommendation. These harder cases therefore provide a natural target for future model development, including post-training, reinforcement learning, and synthetic data generation that reward synthesis, completeness, trade-off analysis, and judgment under uncertainty. 

Subsequent work can pair BusinessCaseBench with interactive protocols, organizational deployment studies, and direct human baselines to connect measured capability to institutional outcomes.

\section{Methods}\label{methods}

\subsection{Benchmark construction}\label{benchmark-construction}

BusinessCaseBench was constructed from licensed business school case PDFs and their paired instructor case solution PDFs, distributed by publishers under restricted use agreements; see \sisecref{supp:released-withheld-artifacts} for the list of released and withheld artifacts. All LLM usage on these materials was conducted under Zero Data Retention (ZDR) policies. Each evaluated instance associates a case narrative with an exam-style question prompt, a reference solution drawn directly from the instructor case solution, an equally-weighted checklist rubric, question-type metadata, and O*NET work-activity tags. The dataset schema is described in \sisecref{supp:final-dataset-schema}, and the compositional breakdown of BusinessCaseBench questions is reported in \sisecref{supp:composition-tables}.

All PDFs were converted to linearized markdown using OCR from the olmOCR toolkit \citep{poznanski_olmocr_2025}, with the \texttt{allenai/olmOCR-2-7B-1025-FP8} vision-language model served via vLLM \citep{kwon_efficient_2023} applied in batch over the full corpus and followed by an LLM-assisted cleaning pass on the resulting text. We audited verbatim exposure of these materials in large web-crawled pre-training corpora using InfiniGram \citep{Liu2024InfiniGram} and found no evidence of contamination in commonly used datasets. \sisecref{supp:contamination-evidence} reports the audit protocol and results for C4 \citep{raffel2020exploring}, The Pile \citep{pile}, RedPajama \citep{together2023redpajama}, Dolma \citep{soldaini-etal-2024-dolma}, and DCLM \citep{li2025datacomplmsearchgenerationtraining}. Prior to question construction, each case was independently labeled as depicting a fictional organization or a named real focal organization.

Questions and reference solutions were extracted solely from material appearing explicitly in the instructor case solution, not inferred from the case narrative, and any instance without an explicit reference solution in the instructor case solution was excluded. Retained instances were standardized through LLM rewrite passes into self-contained exam-style question prompts, and an automated quality gate discarded any extraction that failed basic fidelity checks on the question--reference-solution triple. We further annotated each instance for numerical versus non-numerical content, subjective versus objective framing, business discipline, and O*NET taxonomy. Although the construction pipeline drew on LLM assistance from question extraction through metadata annotation (\texttt{google/gemini-2.5-pro} and \texttt{google/gemini-2.5-flash}, \citep{comanici_gemini_2025}), every retained question, metadata, and reference solution underwent manual review and inspection for quality before inclusion.

For each instance, an equally-weighted checklist rubric was constructed from the exam-style question prompt and the reference solution in the instructor case solution, following established best practices for rubric-guided automated evaluation \citep{min_factscore_2023,lin_wildbench_2024,wei_rocketeval_2025}, with binary credit awarded per criterion and every criterion carrying equal weight for assigning partial credit. Every question was additionally mapped to O*NET Work Activity (WA), Intermediate Work Activity (IWA), and Detailed Work Activity (DWA) labels, with counts by level reported in \sisecref{supp:composition-tables}. Construction and evaluation workflows were orchestrated with DataDreamer \citep{patel_datadreamer_2024} for reproducible LLM pipelines. The evaluation pipeline code is released alongside the paper (see Data availability).

\subsection{Frontier AI models and their evaluation}\label{large-language-models-and-their-evaluation}

The primary cross-model analysis compares three frontier AI models, OpenAI GPT-5.4 \citep{singh_openai_2026}, Anthropic Claude Sonnet~4.6 \citep{anthropic_system_2025}, and Google Gemini~3 Flash Preview \citep{deepmind_gemini_2025}. \sisecref{supp:model-evaluation-protocol} lists the API identifiers and sampling settings for each. To trace capability development within a single model family, we additionally evaluated four successive OpenAI releases spanning roughly two years, from GPT-4 Turbo through GPT-4.1, GPT-5, and GPT-5.4 \citep{openai_gpt-4_2024,singh_openai_2026}, on the same fixed question set.

Each solver model received the complete case narrative and exam-style question prompt in a single turn, with no tool access, retrieval augmentation, or multi-turn clarification; the inference prompt template, provider settings, and token limits appear in \sisecref{supp:model-evaluation-protocol}. Grading was conducted independently of generation by a fixed LLM-as-judge \citep{zheng_judging_2023} held constant across all solver models. We use the smaller, faster, and efficient Gemini~2.5 Flash (\texttt{google/gemini-2.5-flash}) model \citep{comanici_gemini_2025} for the judging task as is typically standard in LLM-as-judge setups. The judge evaluates each attempted solution criterion by criterion, awarding binary credit (0 or 1 point) against the checklist rubric and reference solution, following the protocol in \sisecref{supp:judge-grading-prompt}. All seven models were evaluated on the same 615 questions, with only the solver varying across runs. To assess whether the choice of an LLM judge model reshapes comparative conclusions, we regraded answers from Claude Sonnet~4.6, GPT-5.4, and Gemini~3 Flash Preview with three family-matched small judge models (Gemini~2.5 Flash, Claude Haiku~4.5, and GPT-5-mini) on all $n=615$ questions. We measured agreement in two complementary ways. Rank-order concordance used Kendall's coefficient of concordance ($W$) and pairwise Kendall's $\tau$ on mean scores across the three solvers. Relative-score concordance used pairwise Pearson correlations of each judge's scores over those same solvers, capturing whether judges agree on how far models sit apart while remaining invariant to absolute score level. All three judges produced the same overall ranking ($W=1.0$; all pairwise $\tau=1.0$), and relative score differences likewise agreed almost exactly (pairwise $r\in[0.999,1.000]$). At the question level, agreement was only moderate (mean pairwise $\tau\approx0.56$; mean pairwise $r\approx0.61$; identical ranking on 35\% of questions), as expected from item-level noise, yet these disagreements left the overall rank ordering unchanged. We therefore conclude that the choice of LLM judge model does not significantly reshape comparative conclusions on BusinessCaseBench.

\subsection{Human annotation interface and blinded validation protocol}\label{human-annotation-interface-and-blinded-validation-protocol}

To validate automated rubrics and grading scores against human judgments of answer quality, three annotators with business school grading experience (Annotators~A, B, and~C) completed a blinded evaluation protocol on a stratified random sample. Recruitment and training procedures are described in \sisecref{supp:annotator-recruitment-training}. The workflow was implemented in a custom web interface, shown in \sifigref{fig:annotation_interface}, with staged unblinding that ensured human rubrics and scores were not shaped by automated outputs; see \sisecref{supp:interface-protocol} for the full protocol.

For each assigned question, annotators first judged whether the exam-style question prompt and reference solution were usable and well-formed to further independently validate the quality of the extracted questions and solutions from the case PDFs. Then, annotators independently authored their own checklist rubric and graded the model-generated attempted solution before viewing the automated rubric or grade. Subsequent stages asked annotators to evaluate the automated rubric and grading and to compare both layers directly to their own.

Grading open-ended case responses is necessarily somewhat subjective as different instructors and graders may disagree on acceptance criteria and partial credit thresholds. In this blinded protocol, annotators wrote independent rubrics and often assigned different partial credit scores to the same model response. Given human annotators do not agree upon a single score, we do not expect automated scores to match any single human score exactly. Validation statistics are reported on ten assignments in which all three annotators judged the same case, question, and model response. The results support that the automatically generated rubrics and the LLM-as-judge procedure are directionally consistent with expert judgment and suitable for measuring relative progress across models, disciplines, and time: automated Standard scores correlated with human partial credit scores at Spearman $\rho = 0.54$; annotators rated 100\% of the automatically generated rubrics acceptable or mainly acceptable and 96\% of the LLM-as-judge Standard scores acceptable or mainly acceptable; and in 74\% of the assignments they preferred the automated score or judged it as equivalent to their own. Further details and agreement statistics appear in \sisecref{supp:agreement-results}.

\subsection{Reporting of evaluation metrics and statistics}\label{reporting-of-evaluation-metrics-and-statistics}

We report two primary outcome metrics, formalized as follows. Let model $m$ answer question $j$, which has $k_j$ equally-weighted rubric criteria, and let $c_{mji}\in\{0,1\}$ denote whether criterion $i$ is satisfied under LLM-as-judge grading. The per-question \emph{Standard score} is
\begin{equation}
s_{mj}=\frac{1}{k_j}\sum_{i=1}^{k_j} c_{mji},
\label{eq:standard-question-score}
\end{equation}
a normalized value in $[0,1]$ (reported as a score percentage in figures). The model-level Standard score is the unweighted mean over the $N=615$ benchmark questions,
\begin{equation}
\bar{s}_m=\frac{1}{N}\sum_{j=1}^{N} s_{mj}.
\label{eq:standard-model-mean}
\end{equation}
\emph{Complete Answer scoring} marks whether every criterion is satisfied on a question. Define the indicator
\begin{equation}
a_{mj}=\mathbbm{1}[s_{mj}=1],
\label{eq:complete-answer-indicator}
\end{equation}
and the \emph{Complete Answer score} as its mean over questions,
\begin{equation}
\bar{a}_m=\frac{1}{N}\sum_{j=1}^{N} a_{mj}.
\label{eq:complete-answer-score}
\end{equation}
The same definitions apply when reporting means over a subset of questions within a discipline, question-type strata, or per-IWA. We report full leaderboards and decompositions of results in \sisecref{supp:extended-results}. Uncertainty, when provided, is quantified through nonparametric bootstrap resampling over questions, drawing $B = 500$ resamples with replacement for each metric and reporting percentile 95\% confidence intervals.

\section{Acknowledgments}
The authors thank Mark Yatskar, Daniel Rock, Eric Horvitz, and Mitch Weiss for early discussions during the conceptualization of this work. The authors thank the anonymous reviewers for their valuable suggestions.
\section{Competing interest}
The authors declare that they have no competing interests.
\section{Supplementary material}
Supplementary material is available below.

\section{Funding}
Ramayya Krishnan's work was supported in part by NIST Federal Award ID 60NANB24D231, Carnegie Mellon University’s AI Measurement Science and Engineering Center (AIMSEC), the CMU ACE-AI initiative funded by Accenture, and a research gift from Tata Consultancy Services.

\section{Author contributions}

Conceptualization: A.P., K.H., R.K., and C.C.B.; Data curation: A.P. and K.H.; Methodology, experiments, and analysis: A.P.; Manuscript writing: A.P.; Manuscript review and editing: A.P., K.H., R.K., C.C.B., and K.L.; Resources: K.H., R.K., and K.L.
\section{Data availability}

All data and code (where license terms allow us to share the data) to reproduce the results in this article have been deposited to Zenodo at \url{https://doi.org/10.5281/zenodo.20211020}.

\bibliographystyle{abbrvnat}
\bibliography{reference}

\clearpage
\onecolumn
\setcounter{section}{0}
\setcounter{subsection}{0}
\renewcommand{\thesection}{\Alph{section}}
\renewcommand{\thesubsection}{\thesection\arabic{subsection}}
\makeatletter
\@ifundefined{theHsection}{}{\renewcommand{\theHsection}{supp.\Alph{section}}}
\@ifundefined{theHsubsection}{}{\renewcommand{\theHsubsection}{supp.\Alph{section}.\arabic{subsection}}}
\makeatother

\section*{Supplementary material}
\addcontentsline{toc}{section}{Supplementary material}

Supplementary Information for ``\papertitle'' providing information about additional results, reproducibility details, and experimental details.

\section*{Table of contents}
\noindent\hyperref[supp:benchmark-construction]{A\quad Benchmark construction}\dotfill\pageref{supp:benchmark-construction}\par
\noindent\hspace*{1.5em}\hyperref[supp:final-dataset-schema]{A1\quad Final dataset schema}\dotfill\pageref{supp:final-dataset-schema}\par
\noindent\hspace*{1.5em}\hyperref[supp:composition-tables]{A2\quad Composition tables}\dotfill\pageref{supp:composition-tables}\par
\noindent\hyperref[supp:model-evaluation-protocol]{B\quad Model evaluation protocol}\dotfill\pageref{supp:model-evaluation-protocol}\par
\noindent\hspace*{1.5em}\hyperref[supp:evaluated-models]{B1\quad Evaluated models}\dotfill\pageref{supp:evaluated-models}\par
\noindent\hspace*{1.5em}\hyperref[supp:inference-prompt-template]{B2\quad Inference prompt template}\dotfill\pageref{supp:inference-prompt-template}\par
\noindent\hspace*{1.5em}\hyperref[supp:judge-grading-prompt]{B3\quad LLM-as-judge grading prompt}\dotfill\pageref{supp:judge-grading-prompt}\par
\noindent\hyperref[supp:human-annotation-validation]{C\quad Human annotation and blinded validation}\dotfill\pageref{supp:human-annotation-validation}\par
\noindent\hspace*{1.5em}\hyperref[supp:interface-protocol]{C1\quad Interface and protocol}\dotfill\pageref{supp:interface-protocol}\par
\noindent\hspace*{1.5em}\hyperref[supp:annotator-recruitment-training]{C2\quad Annotator recruitment and training}\dotfill\pageref{supp:annotator-recruitment-training}\par
\noindent\hspace*{1.5em}\hyperref[supp:agreement-results]{C3\quad Rubric and grading validation and agreement results}\dotfill\pageref{supp:agreement-results}\par
\noindent\hyperref[supp:extended-results]{D\quad Extended results}\dotfill\pageref{supp:extended-results}\par
\noindent\hspace*{1.5em}\hyperref[supp:full-leaderboards]{D1\quad Full leaderboards under both scoring regimes}\dotfill\pageref{supp:full-leaderboards}\par
\noindent\hspace*{1.5em}\hyperref[supp:discipline-model-matrix]{D2\quad Per-discipline by per-model score matrix}\dotfill\pageref{supp:discipline-model-matrix}\par
\noindent\hspace*{1.5em}\hyperref[supp:openai-trajectory-discipline]{D3\quad OpenAI performance trajectory by discipline}\dotfill\pageref{supp:openai-trajectory-discipline}\par
\noindent\hspace*{1.5em}\hyperref[supp:question-type-strata]{D4\quad Scoring by question-type strata}\dotfill\pageref{supp:question-type-strata}\par
\noindent\hspace*{1.5em}\hyperref[supp:fictional-real-delta]{D5\quad Fictional-vs-real performance delta by discipline}\dotfill\pageref{supp:fictional-real-delta}\par
\noindent\hspace*{1.5em}\hyperref[supp:metadata-regression]{D6\quad Score variation by question type, discipline, and case}\dotfill\pageref{supp:metadata-regression}\par
\noindent\hspace*{1.5em}\hyperref[supp:cross-model-complementarity]{D7\quad Cross-model complementarity}\dotfill\pageref{supp:cross-model-complementarity}\par
\noindent\hspace*{1.5em}\hyperref[supp:extended-onet-taxonomy-results]{D8\quad Extended O*NET taxonomy results}\dotfill\pageref{supp:extended-onet-taxonomy-results}\par
\noindent\hyperref[supp:validity-limitations-reproducibility]{E\quad Validity, limitations, and reproducibility}\dotfill\pageref{supp:validity-limitations-reproducibility}\par
\noindent\hspace*{1.5em}\hyperref[supp:contamination-evidence]{E1\quad Contamination evidence}\dotfill\pageref{supp:contamination-evidence}\par
\noindent\hspace*{1.5em}\hyperref[supp:scope-of-coverage]{E2\quad Scope of coverage}\dotfill\pageref{supp:scope-of-coverage}\par
\noindent\hspace*{1.5em}\hyperref[supp:released-withheld-artifacts]{E3\quad Released vs. withheld artifacts and access pathway}\dotfill\pageref{supp:released-withheld-artifacts}\par

\clearpage
\refstepcounter{section}\label{supp:benchmark-construction}
\section*{\thesection\quad Benchmark construction}
\setcounter{subsection}{0}

\refstepcounter{subsection}\label{supp:final-dataset-schema}
\subsection*{\thesubsection\quad Final dataset schema}

Each evaluated item is one JSON record: case metadata, full case and instructor case solution text, a standalone exam-style \texttt{question}, a \texttt{solution} derived from instructor case solution, a checklist \texttt{grading\_rubric}, question-type flags, and O*NET work-activity tags. The illustrative record below is a representative example of the format; proprietary text fields are shown as ellipses.

\begin{verbatim}
{
  "case_name": "ILLUSTRATIVE-001",
  "case_title": "Acme Corp: European Expansion",
  "case_summary": "In late 2023, Acme Corp---a $480M U.S. manufacturer---weighed entering the European market...",
  "case_clean_text": "...",
  "instructor_case_solution_clean_text": "...",
  "question": "Provide an analysis of Acme Corp's strategic position and make a recommendation on...",
  "solution": "An analysis should size the European TAM at roughly 11B EUR and note that three incumbents...",
  "task_description": "Evaluate a business's market entry and channel development strategy.",
  "numerical": false,
  "subjective": true,
  "grading_rubric": [
    "Answer identifies European market size (TAM) correctly.",
    "Answer cites competitive threat from incumbent concentration.",
    "Answer computes or discusses break-even economics for entry.",
    "Answer notes regulatory or CE-mark compliance risk.",
    "Answer states a clear enter-or-defer recommendation for 2024.",
    "..."
  ],
  "discipline": "Strategy",
  "work_activity": "Making Decisions and Solving Problems",
  "work_activity_id": "4.A.2.b.1",
  "intermediate_work_activity": "Advise others on business or operational matters.",
  "intermediate_work_activity_id": "4.A.4.b.6.I05",
  "detailed_work_activity": "Measure effectiveness of business strategies or practices.",
  "detailed_work_activity_id": "4.A.2.a.1.I02.D03"
}
\end{verbatim}

\refstepcounter{subsection}\label{supp:composition-tables}
\subsection*{\thesubsection\quad Composition tables}

The following tables summarize the composition of the benchmark used in our analysis. They report the overall coverage and break down the questions in the dataset by strata, discipline, and by the O*NET taxonomy of work activities, providing a compact view of the distribution of questions in the benchmark.


\begin{table}[H]
\caption{Benchmark composition: overall question-level metadata.}\label{tab:overall-metadata}%
\begin{tabular*}{\columnwidth}{@{\extracolsep\fill}lc@{\extracolsep\fill}}
\hline
Strata & Questions \\
\hline
All questions & 615 \\
Fictional case & 245 \\
Real case & 370 \\
Numerical & 184 \\
Non-numerical & 431 \\
Subjective & 339 \\
Objective & 276 \\
\hline
\end{tabular*}
\end{table}


\begin{table}[H]
\caption{Benchmark composition by business discipline. Non-num.\ = non-numerical.}\label{tab:by-discipline}%
\begin{tabular*}{\textwidth}{@{\extracolsep\fill}lrrrrrrrr@{\extracolsep\fill}}
\hline
Discipline & $n$ & Fictional & Real & Numerical & Non-num. & Subjective & Objective \\
\hline
Strategy & 110 & 19 & 91 & 20 & 90 & 77 & 33 \\
Leadership \& Organizational Behavior & 102 & 52 & 50 & 1 & 101 & 81 & 21 \\
Finance & 87 & 32 & 55 & 64 & 23 & 28 & 59 \\
Accounting & 58 & 39 & 19 & 45 & 13 & 9 & 49 \\
Business Ethics & 43 & 30 & 13 & 2 & 41 & 36 & 7 \\
General Management & 34 & 8 & 26 & 4 & 30 & 29 & 5 \\
Operations \& Service Management & 31 & 7 & 24 & 7 & 24 & 12 & 19 \\
Economics & 24 & 1 & 23 & 9 & 15 & 1 & 23 \\
Human Resource Management & 22 & 15 & 7 & 2 & 20 & 16 & 6 \\
Marketing \& Sales & 22 & 7 & 15 & 3 & 19 & 16 & 6 \\
Decision Analysis & 21 & 18 & 3 & 19 & 2 & 3 & 18 \\
Entrepreneurship \& Innovation & 17 & 4 & 13 & 3 & 14 & 11 & 6 \\
Information Technology & 12 & 2 & 10 & 0 & 12 & 1 & 11 \\
Business \& Government Relations & 12 & 1 & 11 & 0 & 12 & 7 & 5 \\
Negotiation & 8 & 2 & 6 & 3 & 5 & 5 & 3 \\
International Business & 7 & 5 & 2 & 1 & 6 & 3 & 4 \\
Management Communications & 4 & 3 & 1 & 0 & 4 & 3 & 1 \\
Social Enterprise & 1 & 0 & 1 & 1 & 0 & 1 & 0 \\
\hline
\textbf{Total} & \textbf{615} & \textbf{245} & \textbf{370} & \textbf{184} & \textbf{431} & \textbf{339} & \textbf{276} \\
\hline
\end{tabular*}
\end{table}


\begin{longtable}{@{}p{0.68\textwidth}lc@{}}
\caption{Benchmark composition by O*NET Work Activity (WA). Rows are ordered by question count descending.}\label{tab:by-wa}\\
\hline
Work Activity & WA ID & $n$ \\
\hline
\endfirsthead
\multicolumn{3}{c}{\small\textit{\sitabref{tab:by-wa} continued.}}\\
\hline
Work Activity & WA ID & $n$ \\
\hline
\endhead
\hline
\endfoot
Analyzing Data or Information & 4.A.2.a.4 & 245 \\
Making Decisions and Solving Problems & 4.A.2.b.1 & 108 \\
Developing Objectives and Strategies & 4.A.2.b.4 & 89 \\
Judging the Qualities of Things, Services, or People & 4.A.2.a.1 & 28 \\
Provide Consultation and Advice to Others & 4.A.4.b.6 & 27 \\
Interpreting the Meaning of Information for Others & 4.A.4.a.1 & 24 \\
Processing Information & 4.A.2.a.2 & 19 \\
Selling or Influencing Others & 4.A.4.a.6 & 15 \\
Estimating the Quantifiable Characteristics of Products, Events, or Information & 4.A.1.b.3 & 13 \\
Evaluating Information to Determine Compliance with Standards & 4.A.2.a.3 & 9 \\
Thinking Creatively & 4.A.2.b.2 & 8 \\
Getting Information & 4.A.1.a.1 & 6 \\
Organizing, Planning, and Prioritizing Work & 4.A.2.b.6 & 5 \\
Identifying Objects, Actions, and Events & 4.A.1.b.1 & 4 \\
Coaching and Developing Others & 4.A.4.b.5 & 4 \\
Documenting/Recording Information & 4.A.3.b.6 & 2 \\
Training and Teaching Others & 4.A.4.b.3 & 2 \\
Monitor Processes, Materials, or Surroundings & 4.A.1.a.2 & 1 \\
Interacting With Computers & 4.A.3.b.1 & 1 \\
Resolving Conflicts and Negotiating with Others & 4.A.4.a.7 & 1 \\
Staffing Organizational Units & 4.A.4.c.2 & 1 \\
Updating and Using Relevant Knowledge & 4.A.2.b.3 & 1 \\
Performing Administrative Activities & 4.A.4.c.1 & 1 \\
Communicating with Persons Outside Organization & 4.A.4.a.3 & 1 \\
\end{longtable}


\begin{longtable}{@{}p{0.62\textwidth}lc@{}}
\caption{Benchmark composition by O*NET Intermediate Work Activity (IWA). Rows ordered by question count descending.}\label{tab:by-iwa}\\
\hline
Intermediate Work Activity & IWA ID & $n$ \\
\hline
\endfirsthead
\multicolumn{3}{c}{\small\textit{\sitabref{tab:by-iwa} continued.}}\\
\hline
Intermediate Work Activity & IWA ID & $n$ \\
\hline
\endhead
\hline
\endfoot
Analyze business or financial data. & 4.A.2.a.4.I11 & 117 \\
Advise others on business or operational matters. & 4.A.4.b.6.I05 & 88 \\
Evaluate programs, practices, or processes. & 4.A.2.a.1.I02 & 48 \\
Analyze business or financial risks. & 4.A.2.a.4.I03 & 44 \\
Research organizational behavior, processes, or performance. & 4.A.1.a.1.I09 & 34 \\
Develop financial or business plans. & 4.A.2.b.2.I09 & 33 \\
Investigate organizational or operational problems. & 4.A.1.a.1.I22 & 31 \\
Develop organizational policies, systems, or processes. & 4.A.2.b.4.I01 & 27 \\
Analyze market or industry conditions. & 4.A.2.a.4.I02 & 26 \\
Develop business or marketing plans. & 4.A.2.b.2.I03 & 20 \\
Assess characteristics or impacts of regulations or policies. & 4.A.2.a.4.I09 & 13 \\
Analyze data to improve operations. & 4.A.2.a.4.I07 & 13 \\
Develop organizational or program goals or objectives. & 4.A.2.b.2.I27 & 13 \\
Prepare financial documents, reports, or budgets. & 4.A.3.b.6.I01 & 9 \\
Calculate financial data. & 4.A.1.b.3.I03 & 8 \\
Advise others on financial matters. & 4.A.4.b.6.I11 & 6 \\
Evaluate the characteristics, usefulness, or performance of products or technologies. & 4.A.2.a.1.I07 & 6 \\
Analyze scientific or applied data using mathematical principles. & 4.A.2.a.4.I04 & 6 \\
Examine financial activities, operations, or systems. & 4.A.2.a.3.I03 & 5 \\
Resolve personnel or operational problems. & 4.A.4.a.7.I03 & 5 \\
Develop contingency or emergency response plans. & 4.A.2.b.2.I19 & 5 \\
Identify business or organizational opportunities. & 4.A.1.b.1.I02 & 4 \\
Evaluate personnel capabilities or performance. & 4.A.2.a.1.I04 & 4 \\
Develop sustainable organizational or business policies or practices. & 4.A.2.b.2.I20 & 4 \\
Develop models of systems, processes, or products. & 4.A.2.b.2.I21 & 4 \\
Explain financial information. & 4.A.4.a.1.I04 & 3 \\
Evaluate condition of financial assets, property, or other resources. & 4.A.2.a.1.I09 & 3 \\
Reconcile financial data. & 4.A.2.a.2.I04 & 3 \\
Evaluate project feasibility. & 4.A.2.a.1.I10 & 3 \\
Determine operational methods or procedures. & 4.A.2.b.1.I04 & 2 \\
Maintain sales or financial records. & 4.A.3.b.6.I10 & 2 \\
Develop research plans or methodologies. & 4.A.2.b.2.I23 & 2 \\
Plan work activities. & 4.A.2.b.6.I02 & 2 \\
Research laws, precedents, or other legal data. & 4.A.2.a.4.I08 & 1 \\
Develop systems or practices to mitigate or resolve environmental problems. & 4.A.2.b.2.I16 & 1 \\
Present research or technical information. & 4.A.3.b.6.I03 & 1 \\
Examine materials or documentation for accuracy or compliance. & 4.A.2.a.3.I01 & 1 \\
Research historical or social issues. & 4.A.1.a.1.I18 & 1 \\
Prepare reports of operational or procedural activities. & 4.A.3.b.6.I15 & 1 \\
Monitor external affairs, trends, or events. & 4.A.1.a.2.I08 & 1 \\
Investigate criminal or legal matters. & 4.A.1.a.1.I03 & 1 \\
Develop professional relationships or networks. & 4.A.4.a.4.I01 & 1 \\
Analyze environmental or geospatial data. & 4.A.2.a.4.I01 & 1 \\
Evaluate production inputs or outputs. & 4.A.2.a.1.I05 & 1 \\
Estimate project development or operational costs. & 4.A.1.b.3.I02 & 1 \\
Collect data about consumer needs or opinions. & 4.A.1.a.1.I14 & 1 \\
Develop marketing or promotional materials. & 4.A.2.b.2.I12 & 1 \\
Monitor operations to ensure adequate performance. & 4.A.1.a.2.I02 & 1 \\
Explain regulations, policies, or procedures. & 4.A.4.a.1.I02 & 1 \\
Evaluate the quality or accuracy of data. & 4.A.2.a.2.I01 & 1 \\
Process digital or online data. & 4.A.3.b.1.I06 & 1 \\
Prepare legal or regulatory documents. & 4.A.3.b.6.I14 & 1 \\
Monitor individual behavior or performance. & 4.A.1.a.2.I06 & 1 \\
Gather data about operational or development activities. & 4.A.1.a.1.I06 & 1 \\
Prepare proposals or grant applications. & 4.A.3.b.6.I07 & 1 \\
\end{longtable}


\begin{longtable}{@{}p{0.58\textwidth}lc@{}}
\caption{Benchmark composition by O*NET Detailed Work Activity (DWA). Rows ordered by question count descending.}\label{tab:by-dwa}\\
\hline
Detailed Work Activity & DWA ID & $n$ \\
\hline
\endfirsthead
\multicolumn{3}{c}{\small\textit{\sitabref{tab:by-dwa} continued.}}\\
\hline
Detailed Work Activity & DWA ID & $n$ \\
\hline
\endhead
\hline
\endfoot
Analyze business or financial data. & 4.A.2.a.4.I11.D04 & 57 \\
Develop operating strategies, plans, or procedures. & 4.A.2.b.4.I01.D02 & 37 \\
Advise others on business or operational matters. & 4.A.4.b.6.I05.D10 & 33 \\
Analyze financial information. & 4.A.2.a.4.I11.D02 & 29 \\
Measure effectiveness of business strategies or practices. & 4.A.2.a.1.I02.D03 & 27 \\
Analyze operational data to evaluate operations, processes or products. & 4.A.2.a.4.I07.D03 & 27 \\
Develop business or market strategies. & 4.A.2.b.2.I03.D01 & 26 \\
Analyze data to inform operational decisions or activities. & 4.A.2.a.4.I07.D12 & 23 \\
Analyze data to assess operational or project effectiveness. & 4.A.2.a.4.I07.D09 & 19 \\
Analyze market conditions or trends. & 4.A.2.a.4.I02.D04 & 18 \\
Assess risks to business operations. & 4.A.2.a.4.I03.D01 & 15 \\
Analyze risks to minimize losses or damages. & 4.A.2.a.4.I03.D03 & 14 \\
Analyze financial records or reports to determine state of operations. & 4.A.2.a.4.I11.D06 & 13 \\
Develop organizational policies or programs. & 4.A.2.b.4.I01.D01 & 13 \\
Analyze data to identify or resolve operational problems. & 4.A.2.a.4.I07.D14 & 12 \\
Develop financial or business plans. & 4.A.2.b.2.I09.D02 & 12 \\
Analyze costs and benefits of proposed designs or projects. & 4.A.2.a.4.I05.D08 & 10 \\
Apply mathematical models of financial or business conditions. & 4.A.2.a.4.I11.D03 & 10 \\
Develop organizational goals or objectives. & 4.A.2.b.2.I27.D02 & 10 \\
Analyze budgetary or accounting data. & 4.A.2.a.4.I11.D01 & 9 \\
Apply mathematical principles or statistical approaches to solve problems in scientific or applied fields. & 4.A.2.a.4.I04.D01 & 9 \\
Evaluate effectiveness of personnel policies or practices. & 4.A.2.a.1.I02.D04 & 8 \\
Calculate financial data. & 4.A.1.b.3.I03.D01 & 7 \\
Establish business management methods. & 4.A.2.b.4.I01.D05 & 7 \\
Develop marketing plans or strategies. & 4.A.2.b.2.I03.D03 & 7 \\
Develop sustainable business strategies or practices. & 4.A.2.b.2.I20.D02 & 7 \\
Recommend organizational process or policy changes. & 4.A.4.b.6.I05.D07 & 6 \\
Analyze industry trends. & 4.A.2.a.4.I02.D02 & 5 \\
Develop organizational methods or procedures. & 4.A.2.b.4.I01.D04 & 5 \\
Examine financial records to ensure compliance with policies or regulations. & 4.A.2.a.3.I03.D05 & 5 \\
Analyze operational or research data. & 4.A.2.a.4.I07.D02 & 5 \\
Analyze data to identify trends or relationships among variables. & 4.A.2.a.4.I04.D02 & 5 \\
Analyze financial records to improve budgeting or planning. & 4.A.2.a.4.I11.D05 & 4 \\
Analyze forecasting data to improve business decisions. & 4.A.2.a.4.I07.D08 & 4 \\
Evaluate civic projects or public policies. & 4.A.2.a.1.I02.D07 & 4 \\
Determine causes of operational problems or failures. & 4.A.2.b.1.I02.D01 & 4 \\
Analyze market or customer related data. & 4.A.2.a.4.I02.D06 & 4 \\
Advise others on ways to improve processes or products. & 4.A.4.b.6.I05.D11 & 4 \\
Prepare financial documents, reports, or budgets. & 4.A.3.b.6.I01.D02 & 4 \\
Evaluate applicable laws and regulations to determine impact on organizational activities. & 4.A.2.a.4.I09.D04 & 4 \\
Calculate data to inform organizational operations. & 4.A.2.a.4.I07.D01 & 3 \\
Develop sustainable organizational policies or practices. & 4.A.2.b.2.I20.D03 & 3 \\
Interpret research or operational data. & 4.A.2.a.4.I07.D05 & 3 \\
Develop plans for programs or services. & 4.A.2.b.2.I27.D03 & 3 \\
Evaluate potential of products, technologies, or resources. & 4.A.2.a.1.I07.D03 & 3 \\
Evaluate program effectiveness. & 4.A.2.a.1.I02.D01 & 3 \\
Develop contingency plans to deal with organizational emergencies. & 4.A.2.b.2.I19.D03 & 3 \\
Maintain financial or account records. & 4.A.3.b.6.I10.D03 & 2 \\
Prepare financial documents. & 4.A.3.b.6.I01.D01 & 2 \\
Maintain knowledge of business operations. & 4.A.2.b.3.I01.D17 & 2 \\
Conduct quantitative failure analyses of operational data. & 4.A.2.a.4.I12.D01 & 2 \\
Conduct research on social issues. & 4.A.1.a.1.I18.D03 & 2 \\
Evaluate project designs to determine adequacy or feasibility. & 4.A.2.a.1.I10.D01 & 2 \\
Negotiate contracts with clients or service providers. & 4.A.4.a.7.I02.D10 & 2 \\
Devise research or testing protocols. & 4.A.2.b.2.I23.D01 & 2 \\
Recommend investments to clients. & 4.A.4.b.6.I11.D02 & 2 \\
Analyze impact of legal or regulatory changes. & 4.A.2.a.4.I09.D03 & 2 \\
Develop emergency response plans or procedures. & 4.A.2.b.2.I19.D04 & 2 \\
Determine pricing or monetary policies. & 4.A.2.b.4.I01.D06 & 2 \\
Monitor financial indicators. & 4.A.1.a.2.I03.D05 & 1 \\
Evaluate employee performance. & 4.A.2.a.1.I04.D04 & 1 \\
Resolve operational performance problems. & 4.A.4.a.7.I03.D02 & 1 \\
Analyze market research data. & 4.A.2.a.4.I02.D08 & 1 \\
Investigate legal issues. & 4.A.1.a.1.I03.D04 & 1 \\
Make decisions in legal cases. & 4.A.2.b.1.I05.D01 & 1 \\
Develop marketing plans or strategies for environmental initiatives. & 4.A.2.b.2.I03.D02 & 1 \\
Develop environmental sustainability plans or projects. & 4.A.2.b.2.I08.D03 & 1 \\
Develop scientific or mathematical models. & 4.A.2.b.2.I26.D03 & 1 \\
Advise others on human resources topics. & 4.A.4.b.6.I05.D08 & 1 \\
Assess product or process usefulness. & 4.A.2.a.1.I07.D05 & 1 \\
Evaluate reports or designs to determine work needs. & 4.A.1.a.1.I02.D12 & 1 \\
Identify investment opportunities or strategies. & 4.A.1.b.1.I02.D03 & 1 \\
Advise others on financial matters. & 4.A.4.b.6.I11.D03 & 1 \\
Assess financial status of clients. & 4.A.2.a.1.I09.D02 & 1 \\
Evaluate the effectiveness of counseling or educational programs. & 4.A.2.a.1.I02.D02 & 1 \\
Establish interpersonal business relationships to facilitate work activities. & 4.A.4.a.4.I01.D04 & 1 \\
Determine the value of goods or services. & 4.A.2.b.1.I01.D02 & 1 \\
Advise others on analytical techniques. & 4.A.4.b.6.I05.D05 & 1 \\
Develop program goals or plans. & 4.A.2.b.2.I27.D01 & 1 \\
Design research studies to obtain scientific information. & 4.A.2.b.2.I23.D04 & 1 \\
Analyze risks related to investments in green technology. & 4.A.2.a.4.I03.D02 & 1 \\
Implement advertising or marketing initiatives. & 4.A.2.b.1.I09.D04 & 1 \\
Prepare analytical reports. & 4.A.3.b.6.I03.D05 & 1 \\
Resolve personnel problems. & 4.A.4.a.7.I03.D03 & 1 \\
Measure environmental characteristics. & 4.A.1.a.2.I09.D03 & 1 \\
Recommend changes or corrective procedures. & 4.A.4.b.6.I05.D12 & 1 \\
Develop detailed project plans. & 4.A.2.b.6.I02.D08 & 1 \\
Analyze data to inform personnel decisions. & 4.A.2.a.4.I07.D11 & 1 \\
Monitor external factors impacting operations. & 4.A.1.a.2.I08.D04 & 1 \\
Develop proposals for current or prospective customers. & 4.A.3.b.6.I07.D01 & 1 \\
Determine operational procedures. & 4.A.2.b.1.I04.D04 & 1 \\
Document organizational or operational procedures. & 4.A.3.b.6.I08.D30 & 1 \\
Identify strategic business investment opportunities. & 4.A.1.b.1.I02.D04 & 1 \\
Communicate with clients about products, procedures, and policies. & 4.A.4.a.1.I02.D06 & 1 \\
Identify opportunities to improve operational efficiency. & 4.A.1.b.1.I02.D06 & 1 \\
Assess the cost effectiveness of products, projects, or services. & 4.A.2.a.1.I07.D01 & 1 \\
Estimate costs of products, services, or materials. & 4.A.2.b.1.I01.D01 & 1 \\
Develop database parameters or specifications. & 4.A.2.b.2.I06.D01 & 1 \\
Prepare data for analysis. & 4.A.3.b.1.I06.D07 & 1 \\
Explain technical product or service information to customers. & 4.A.4.a.1.I01.D05 & 1 \\
Develop operating strategies, plans, or procedures for green or sustainable operations. & 4.A.2.b.2.I20.D01 & 1 \\
Conduct scientific research of organizational behavior or processes. & 4.A.1.a.1.I09.D03 & 1 \\
Research topics in area of expertise. & 4.A.2.b.3.I01.D10 & 1 \\
Analyze jobs using observation, survey, or interview techniques. & 4.A.2.a.4.I07.D10 & 1 \\
Evaluate environmental or sustainability projects. & 4.A.2.a.4.I05.D07 & 1 \\
Recruit personnel. & 4.A.4.c.2.I01.D05 & 1 \\
Forecast economic, political, or social trends. & 4.A.2.a.4.I02.D03 & 1 \\
\end{longtable}

\refstepcounter{section}\label{supp:model-evaluation-protocol}
\section*{\thesection\quad Model evaluation protocol}
\setcounter{subsection}{0}

\refstepcounter{subsection}\label{supp:evaluated-models}
\subsection*{\thesubsection\quad Evaluated models}

The following table lists every evaluated model: provider, API identifier, approximate release date, and inference role. For LLM requests, we use temperature 0; where the API supports it, we enable provider reasoning (approximately \texttt{medium} effort, excluded from the graded response). Each solver completion is capped at 10{,}000 tokens for reasoning and final answer combined.


\begin{table}[H]
\caption{Models evaluated in this study. All evaluations are single-turn with
temperature set to 0. The frontier trio (GPT-5.4, Claude Sonnet~4.6, and
Gemini~3 Flash Preview) are used for primary cross-model comparisons; the four OpenAI
models constitute the within-family generational trajectory. Claude Fable~5 is used for
an after-the-fact Extended Result as a new model release that occurred during the course
of this research; it is not part of the frontier trio.
}\label{tab:models}%
\begin{tabular*}{\textwidth}{@{\extracolsep\fill}llllcc@{\extracolsep\fill}}
\hline
Model name & Provider & API identifier & Approx.\ release & Role & Temp. \\
\hline
GPT-4 Turbo & OpenAI & \texttt{gpt-4-turbo} & Nov 2023 & Trajectory & 0 \\
GPT-4.1 & OpenAI & \texttt{gpt-4.1} & Apr 2025 & Trajectory & 0 \\
GPT-5 & OpenAI & \texttt{gpt-5} & Jan 2026 & Trajectory & 0 \\
GPT-5.4 & OpenAI & \texttt{gpt-5.4} & Mar 2026 & Trajectory + Frontier trio & 0 \\
Claude Sonnet~4.6 & Anthropic & \texttt{claude-sonnet-4.6} & Feb 2026 & Frontier trio & 0 \\
Gemini~3 Flash Preview & Google DeepMind & \texttt{gemini-3-flash-preview} & Dec 2025 & Frontier trio & 0 \\
\hline
Claude Fable~5 & Anthropic & \texttt{claude-fable-5} & June 2026 & Extended Result & 0 \\
\hline
\end{tabular*}
\end{table}

\refstepcounter{subsection}\label{supp:inference-prompt-template}
\subsection*{\thesubsection\quad Inference prompt template}

Each solver model receives the full case text and the benchmark question in a single-turn prompt:

\begin{verbatim}
You are given a business case and a question about the case. You must output
an answer that thoroughly addresses the question and instructions, with full
reasoning and justification for your answer.

The Question:
'''
{question}
'''

The Case:
'''
{case_clean_text}
'''
\end{verbatim}

\refstepcounter{subsection}\label{supp:judge-grading-prompt}
\subsection*{\thesubsection\quad LLM-as-judge grading prompt}

A fixed judge model (\texttt{google/gemini-2.5-flash}) grades each attempted answer at temperature 0 (provider reasoning disabled), with the same 10{,}000-token completion cap. The judge assigns 0 or 1 point per rubric criterion and returns the total score between \texttt{<<} and \texttt{>>} brackets:

\begin{verbatim}
You are an experienced judge of LLM-generated answers. You are given an
LLM-generated answer to a question about a business case and a grading rubric
for the model's answer to that question.

Remember that business cases sometimes involve subjective questions meant to
test critical-thinking and reasoning skills that have no objective correct
answers. Other times, there are objective questions that have a single
correct answer.

You will be provided a rubric of criteria items to grade the LLM-generated
answer against along with an example of a gold standard answer or description
of an answer to the question taken from the instructor case solution for this
business case.

**Your task:** Reason through each rubric criterion and grade the LLM-generated
answer against it. Each criterion is worth 1 point; assign 0 or 1 point per
criterion.

**Output:** Write out your reasoning for each criterion, then return your
final total score for the LLM-generated answer between << and >> brackets.

Question:
'''
{question}
'''

The LLM-Generated Answer:
'''
{model_answer}
'''

The Grading Rubric:
'''
{grading_rubric_list}
'''

Gold Standard Answer:
'''
{solution}
'''

Case Summary:
'''
{case_summary}
'''
\end{verbatim}
\refstepcounter{section}\label{supp:human-annotation-validation}
\section*{\thesection\quad Human annotation and blinded validation}
\setcounter{subsection}{0}

\refstepcounter{subsection}\label{supp:interface-protocol}
\subsection*{\thesubsection\quad Interface and protocol}

Annotators complete a seven-step \emph{Rubric \& Score Validation} workflow in a web interface (\sifigref{fig:annotation_interface}). Staged unblinding keeps human-made rubrics and grades from being biased or shaped by automatic model outputs. Throughout, annotators can read the case PDF, instructor case solution PDF, question, and expert-written reference solution; the attempted answer and automated scores are revealed only when required and after the annotator has created their own rubric and graded the attempted solution against it.

\begin{enumerate}
  \item \textbf{Evaluate question and solution.} Judge whether the benchmark question and instructor reference solution are usable and well-formed (\emph{no} attempted answer; \emph{no} automated rubric or score). \emph{Options:} yes / no on the question; yes / no on the solution; optional brief justification (early exit if either is no).
  \item \textbf{Write an independent rubric.} Author a checklist of grading criteria (\emph{still no} attempted answer; \emph{no} automated rubric or score). \emph{Options:} free-form criterion lines (minimum three).
  \item \textbf{Grade the attempted answer.} The model's response is shown for the first time; mark which human-rubric criteria it satisfies (\emph{still no} automated rubric or score). \emph{Options:} met / not met per criterion (human grading score computed).
  \item \textbf{Evaluate the automated rubric.} The pipeline's checklist rubric is shown. \emph{Options:} acceptable / mainly acceptable / unacceptable; optional justification.
  \item \textbf{Evaluate the automated grading.} The LLM-as-judge score and reasoning are shown. \emph{Options:} acceptable / mainly acceptable / unacceptable; optional justification.
  \item \textbf{Compare rubrics.} Side-by-side view of human and automated checklists. \emph{Options:} A strong alignment / B divergent but valid / C divergent and invalid; optional justification.
  \item \textbf{Compare scores.} Side-by-side view of human and automated grading scores. \emph{Options:} A prefer automated / B equivalent / C prefer human; optional justification.
\end{enumerate}

\begin{Figure}
\centering
\includegraphics[width=0.9\textwidth]{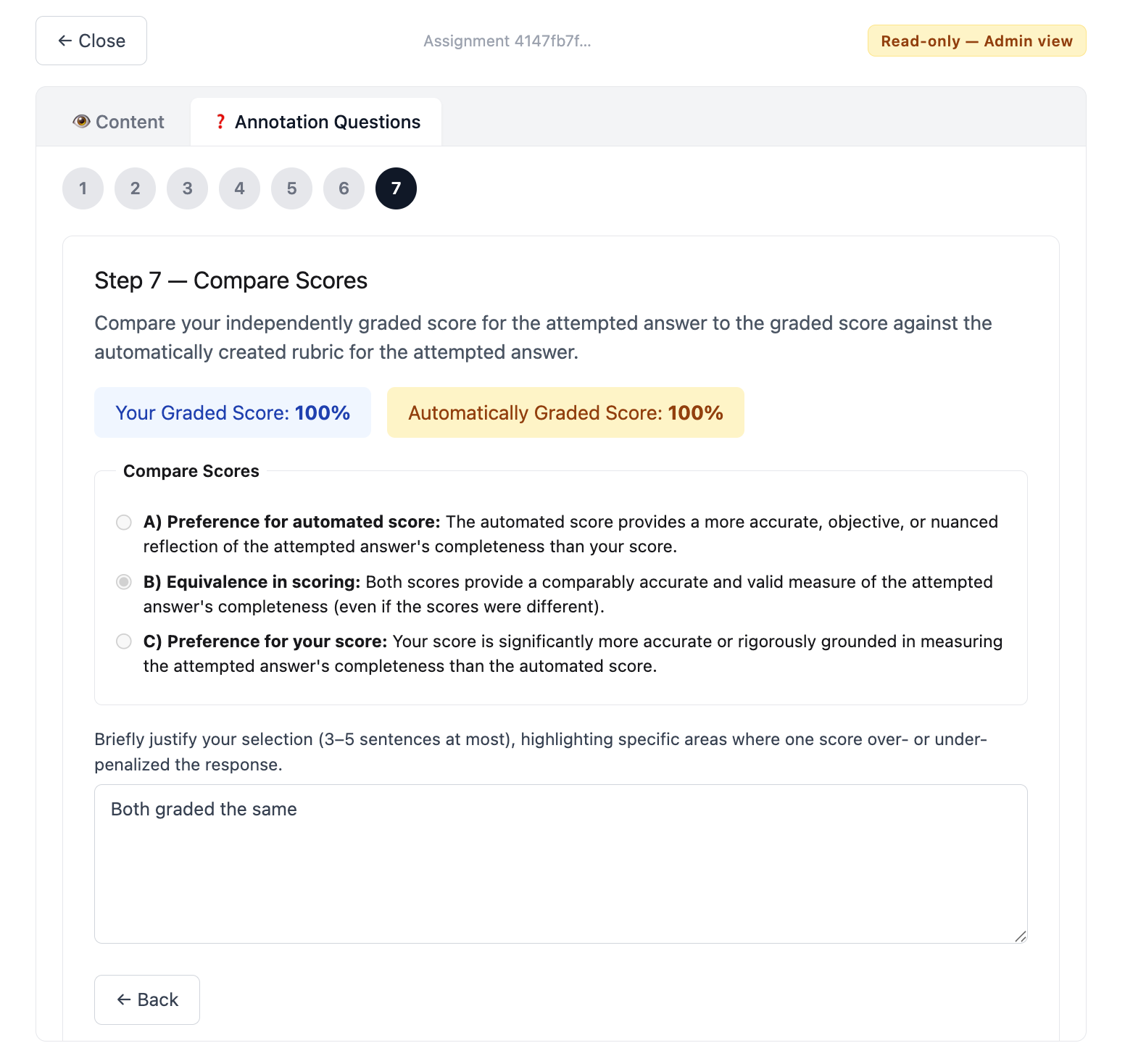}
\caption{Human annotation interface used for blinded validation.}
\label{fig:annotation_interface}
\end{Figure}

\refstepcounter{subsection}\label{supp:annotator-recruitment-training}
\subsection*{\thesubsection\quad Annotator recruitment and training}

Three annotators with training in business school case grading completed the validation layer (denoted Annotators~A, B, and~C). Before scoring live instances, each walked through the interface instructions and a practice assignment covering all seven steps. Task instances were drawn from a stratified random sample of the benchmark. The agreement analysis below focuses on ten annotation assignments in which all three annotators judged the same cases, benchmark questions, and model-generated responses.

\refstepcounter{subsection}\label{supp:agreement-results}
\subsection*{\thesubsection\quad Rubric and grading validation and agreement results}

\sitabref{tab:annotation-validation-summary} summarizes validation metrics across the seven protocol steps. The central question is whether an automatically generated rubric, scored consistently by a fixed LLM judge, is directionally consistent with expert judgment and receives acceptability endorsements from human graders who have independently produced their own rubrics and scores.


\begin{table}[H]
\small
\caption{Summary validation metrics from the blinded human-annotation protocol on the ten annotation assignments (same case, question, and model response) judged independently by Annotators~A, B, and~C.}
\label{tab:annotation-validation-summary}%
\begin{tabular*}{\textwidth}{@{\extracolsep\fill}llccc@{\extracolsep\fill}}
\hline
Step & Validation target & Primary metric & Value & Supporting metric \\
\hline
1 & Question and solution quality & \% yes (question / solution) & 100\% / 100\% & 100\% agreement \\
2 & Independent human rubrics & Mean criteria items per rubric & 10.7 & Mean pairwise difference 3.6 criteria items \\
3 & Human vs.\ automated scores & Spearman $\rho$ & 0.54 & 40\% within 10 percentage points \\
4 & Automated rubric quality & \% acceptable or mainly acceptable & 100\% & 100\% agreement \\
5 & Automated grading quality & \% acceptable or mainly acceptable & 96\% & 90\% agreement \\
6 & Rubric alignment & \% strong alignment (A) or divergent but still valid (B) & 96\% & 90\% agreement \\
7 & Score equivalence & \% prefer automated (A) or equivalent (B) & 74\% & Mean bias $-6.4$ percentage points \\
\hline
\end{tabular*}
\end{table}

In Step~1, all three annotators rated each question and reference solution as usable and well-formed (100\% yes on both dimensions; 100\% agreement). In Step~2, annotators wrote independent rubrics on every question (mean 10.7 criteria items; mean pairwise difference 3.6 criteria items). Variation in rubric lengths across human annotators is expected because grading open-ended cases is somewhat subjective, even when graders work from the same case and reference solution materials.

In Step~3, human partial credit scores correlate with automated Standard scores at Spearman $\rho = 0.54$. For 40\% of annotation assignments, the human and automated graded scores differed by no more than ten percentage points, compared with 46.7\% of inter-annotator score pairs that fell within the same ten-percentage-point band. Thus, automated--human score agreement is close to the level of agreement observed among independent human graders. The automated judge is on average 6.4 percentage points more lenient. Agreement is often close on straightforward items and wider when annotators wrote rubrics of different granularity or applied different partial credit thresholds to the same response.

After viewing the automated outputs, annotators rated the automatically generated rubric acceptable or mainly acceptable on 100\% of completed Step~4 judgments and rated the LLM-as-judge Standard score acceptable or mainly acceptable on 96\% of Step~5 judgments. The one unacceptable grading judgment occurred on a case where human scores ranged from 0\% to 100\% across annotators with different rubrics indicating high disagreement among the annotators. In Step~6, 96\% of rubric comparisons were classified as strongly aligned or divergent but still valid (A or B), and one comparison was rated invalid (C). In Step~7, 74\% of score comparisons preferred the automated Standard score or judged the two scores equivalent (A or B).

These results do not establish exact agreement between automated and human grading scores, and they should not be read that way given the subjectivity visible in Steps~2--3 and~7. They do support a narrower claim that is sufficient for benchmark use: retained questions and instructor-derived reference solutions passed human quality screening; expert graders endorsed the automatically generated rubric and LLM-as-judge Standard scores after working independently; and automated Standard scores correlate with human partial credit scores in the same direction (Spearman $\rho = 0.54$). The validation evidence indicates that the automatic grading procedure is directionally aligned with expert judgment and appropriate for tracking relative progress across models, disciplines, and model generations.

\raggedbottom
\pagebreak

\refstepcounter{section}\label{supp:extended-results}
\section*{\thesection\quad Extended results}
\setcounter{subsection}{0}

\refstepcounter{subsection}\label{supp:full-leaderboards}
\subsection*{\thesubsection\quad Full leaderboards under both scoring regimes}

The tables below report aggregate performance across all 615 questions under Standard scoring and Complete Answer scoring, with bootstrapped 95\% confidence intervals ($B = 500$ resamples).


\begin{table}[H]
\caption{Full aggregate leaderboard under Standard scoring on all 615 benchmark questions. 95\% confidence intervals are bootstrapped
($B = 500$ resamples). Models are sorted by mean score descending.
The four OpenAI models are evaluated on the same fixed question set, enabling
direct within-family comparison; GPT-5.4 also serves as the OpenAI representative in
primary cross-model comparisons. Claude Fable~5 is reported below the rule as an
after-the-fact Extended Result from a new model release during the course of this research;
it is not part of the primary ranking.
}\label{tab:standard-leaderboard}%
\begin{tabular*}{\columnwidth}{@{\extracolsep\fill}lccc@{\extracolsep\fill}}
\hline
Model & Score & 95\% CI lower & 95\% CI upper \\
\hline
\textbf{Claude Sonnet~4.6} & \textbf{88.4\%} & 87.1\% & 89.7\% \\
GPT-5.4 & 87.2\% & 85.8\% & 88.7\% \\
Gemini~3 Flash Preview & 81.6\% & 80.0\% & 83.4\% \\
GPT-5 & 84.9\% & 82.8\% & 86.8\% \\
GPT-4.1 & 80.7\% & 78.8\% & 82.4\% \\
GPT-4 Turbo & 63.9\% & 62.0\% & 65.9\% \\
\hline
Claude Fable~5 & 88.0\% & 86.4\% & 89.5\% \\
\hline
\end{tabular*}
\end{table}


\begin{table}[H]
\caption{Full aggregate leaderboard under Complete Answer scoring on all 615 benchmark questions.
95\% confidence intervals are bootstrapped ($B = 500$ resamples). Models are
sorted by Complete Answer score descending. See \sitabref{tab:standard-leaderboard}
for Standard scoring. Claude Fable~5 is reported below the rule as an
after-the-fact Extended Result from a new model release during the course of this research;
it is not part of the primary ranking.
}\label{tab:complete-leaderboard}%
\begin{tabular*}{\columnwidth}{@{\extracolsep\fill}lccc@{\extracolsep\fill}}
\hline
Model & Score & 95\% CI lower & 95\% CI upper \\
\hline
\textbf{Claude Sonnet~4.6} & \textbf{49.6\%} & 45.9\% & 53.8\% \\
GPT-5.4 & 47.6\% & 43.9\% & 51.4\% \\
Gemini~3 Flash Preview & 32.0\% & 28.6\% & 35.6\% \\
GPT-5 & 46.5\% & 42.4\% & 50.6\% \\
GPT-4.1 & 32.4\% & 28.6\% & 35.9\% \\
GPT-4 Turbo & 13.2\% & 10.7\% & 15.8\% \\
\hline
\textbf{Claude Fable~5} & \textbf{50.9\%} & 47.3\% & 54.8\% \\
\hline
\end{tabular*}
\end{table}

\refstepcounter{subsection}\label{supp:discipline-model-matrix}
\subsection*{\thesubsection\quad Per-discipline by per-model score matrix}

This matrix reports mean Standard scores for every evaluated model within each of the eighteen business disciplines, highlighting where capability is uniform and where models diverge.


\begin{table}[H]
\small
\caption{Per-discipline performance matrix for the frontier trio under Standard scoring and Complete Answer scoring. Mean is the unweighted
average of the three frontier models from the primary experiments only. Disciplines are ordered by ascending mean Standard scoring.
Bold values in the frontier-trio columns indicate the highest score within each scoring block and discipline row among those three models.
Claude Fable~5 appears to the right of each Mean (dotted rule) as an after-the-fact Extended Result from a new model release during the course of this research; it is excluded from the Mean. Fable~5 scores are bolded when strictly higher than all three frontier models in that scoring block.
L\&OB = Leadership \& Organizational Behavior; HRM = Human Resource Management;
E\&I = Entrepreneurship \& Innovation; IT = Information Technology;
B\&GR = Business \& Government Relations.
}\label{tab:discipline-model-matrix}%
\begin{tabular*}{\textwidth}{@{\extracolsep\fill}lrcccc:c|cccc:c@{\extracolsep\fill}}
\hline
 & & \multicolumn{5}{c}{Standard scoring (\%)} & \multicolumn{5}{c}{Complete Answer scoring (\%)} \\
\cline{3-7}\cline{8-12}
Discipline & $n$ & GPT-5.4 & Claude & Gemini & Mean & Fable~5 & GPT-5.4 & Claude & Gemini & Mean & Fable~5 \\
\hline

Marketing \& Sales & 22 & \textbf{82.3} & 80.2 & 77.9 & 80.1 & \textbf{85.2} & \textbf{36.4} & \textbf{36.4} & 22.7 & 31.8 & \textbf{40.9} \\
Finance & 87 & 83.1 & \textbf{83.7} & 74.5 & 80.4 & 82.5 & 41.4 & \textbf{46.0} & 21.8 & 36.4 & 42.5 \\
Operations \& Service Mgmt. & 31 & 83.3 & \textbf{84.4} & 75.2 & 81.0 & 80.3 & 29.0 & \textbf{35.5} & 12.9 & 25.8 & 32.3 \\
Negotiation & 8 & 78.0 & \textbf{88.8} & 77.4 & 81.4 & \textbf{90.5} & 25.0 & \textbf{37.5} & 25.0 & 29.2 & 37.5 \\
Accounting & 58 & 87.2 & \textbf{87.7} & 74.4 & 83.1 & \textbf{88.6} & 48.3 & \textbf{50.0} & 37.9 & 45.4 & \textbf{62.1} \\
General Management & 34 & 83.9 & \textbf{88.6} & 80.5 & 84.4 & 85.0 & \textbf{32.4} & \textbf{32.4} & 14.7 & 26.5 & \textbf{35.3} \\
HRM & 22 & \textbf{88.4} & 87.2 & 79.7 & 85.1 & 86.7 & 45.5 & \textbf{50.0} & 36.4 & 43.9 & 45.5 \\
L\&OB & 102 & 84.7 & \textbf{88.8} & 81.8 & 85.1 & \textbf{89.1} & 46.1 & \textbf{50.0} & 26.5 & 40.8 & 46.1 \\
Business Ethics & 43 & 87.0 & \textbf{90.9} & 86.1 & 88.0 & \textbf{91.4} & \textbf{55.8} & \textbf{55.8} & 46.5 & 52.7 & \textbf{65.1} \\
E\&I & 17 & \textbf{91.0} & 90.7 & 86.8 & 89.5 & \textbf{94.8} & 58.8 & \textbf{70.6} & 35.3 & 54.9 & 52.9 \\
Strategy & 110 & 91.1 & \textbf{91.5} & 85.9 & 89.5 & 87.3 & 44.5 & \textbf{46.4} & 33.6 & 41.5 & \textbf{47.3} \\
Economics & 24 & \textbf{93.7} & 90.7 & 86.8 & 90.4 & \textbf{94.0} & \textbf{70.8} & 50.0 & 37.5 & 52.8 & 62.5 \\
International Business & 7 & \textbf{95.2} & 85.4 & 91.5 & 90.7 & 89.7 & \textbf{71.4} & \textbf{71.4} & 28.6 & 57.1 & 71.4 \\
Decision Analysis & 21 & \textbf{93.9} & 90.2 & 91.4 & 91.9 & 93.6 & \textbf{85.7} & \textbf{85.7} & 76.2 & 82.5 & 81.0 \\
IT & 12 & 88.8 & \textbf{95.8} & 92.5 & 92.4 & \textbf{97.1} & 50.0 & \textbf{75.0} & 58.3 & 61.1 & 75.0 \\
B\&GR & 12 & \textbf{96.1} & 94.2 & 94.8 & 95.0 & \textbf{100.0} & \textbf{83.3} & 58.3 & 66.7 & 69.4 & \textbf{100.0} \\
\hline
\end{tabular*}
\end{table}

\refstepcounter{subsection}\label{supp:openai-trajectory-discipline}
\subsection*{\thesubsection\quad OpenAI performance trajectory by discipline}

The figure tracks the four OpenAI models on the twelve disciplines with at least fifteen questions, under both scoring regimes, to show how generational gains vary by discipline.

\begin{Figure}
\centering
\includegraphics[width=\textwidth]{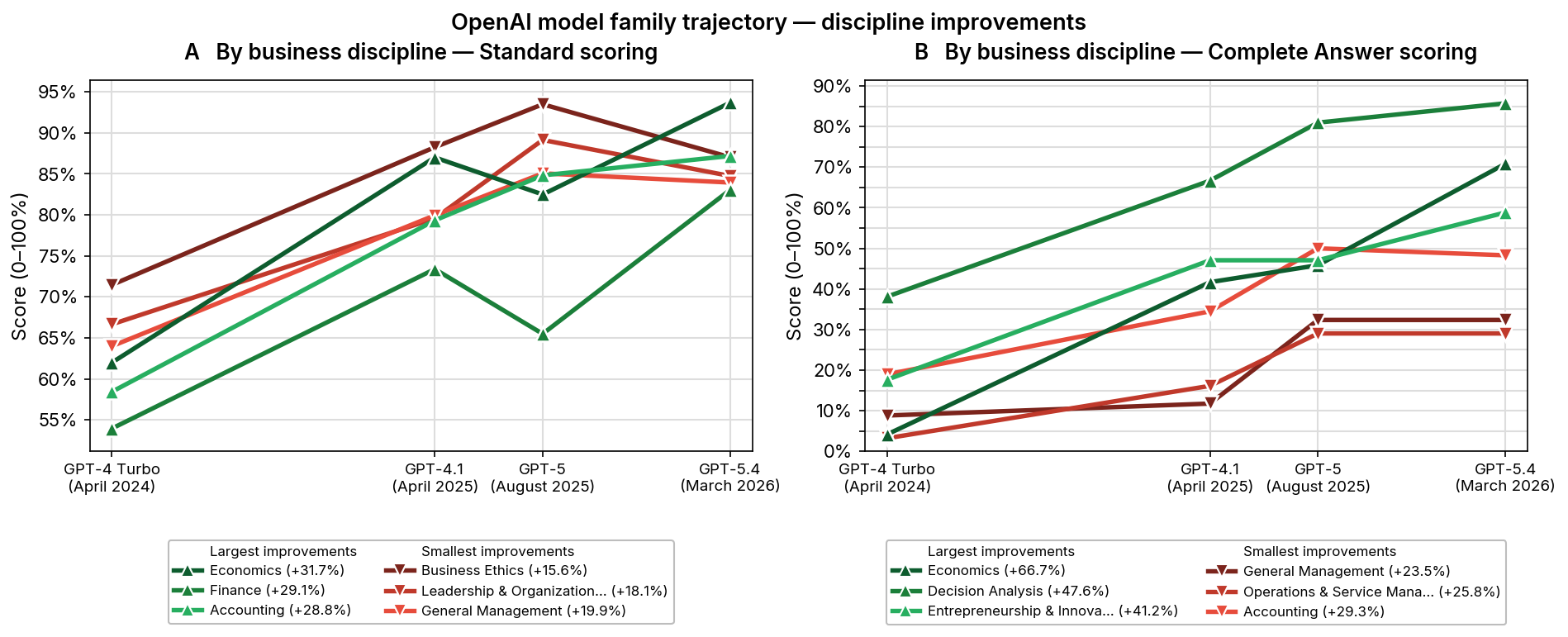}
\caption{Discipline-level OpenAI model family trajectories (twelve disciplines with $n \geq 15$ questions). \textbf{A)} Standard scoring by model; lines highlight the three largest and three smallest discipline-level gains from GPT-4 Turbo to GPT-5.4. \textbf{B)} Scores under Complete Answer scoring by model for the same disciplines. Every tracked discipline improves under both metrics, but gain magnitudes and terminal levels differ.}
\label{fig:openai_trajectory_discipline}
\end{Figure}

\refstepcounter{subsection}\label{supp:question-type-strata}
\subsection*{\thesubsection\quad Scoring by question-type strata}

We report mean frontier performance (under Standard and Complete Answer scoring) by various question-type strata, including whether questions are derived from fictional versus real cases, numerical versus non-numerical questions, and subjective versus objective questions.

\begin{Figure}
\centering
\includegraphics[width=\textwidth]{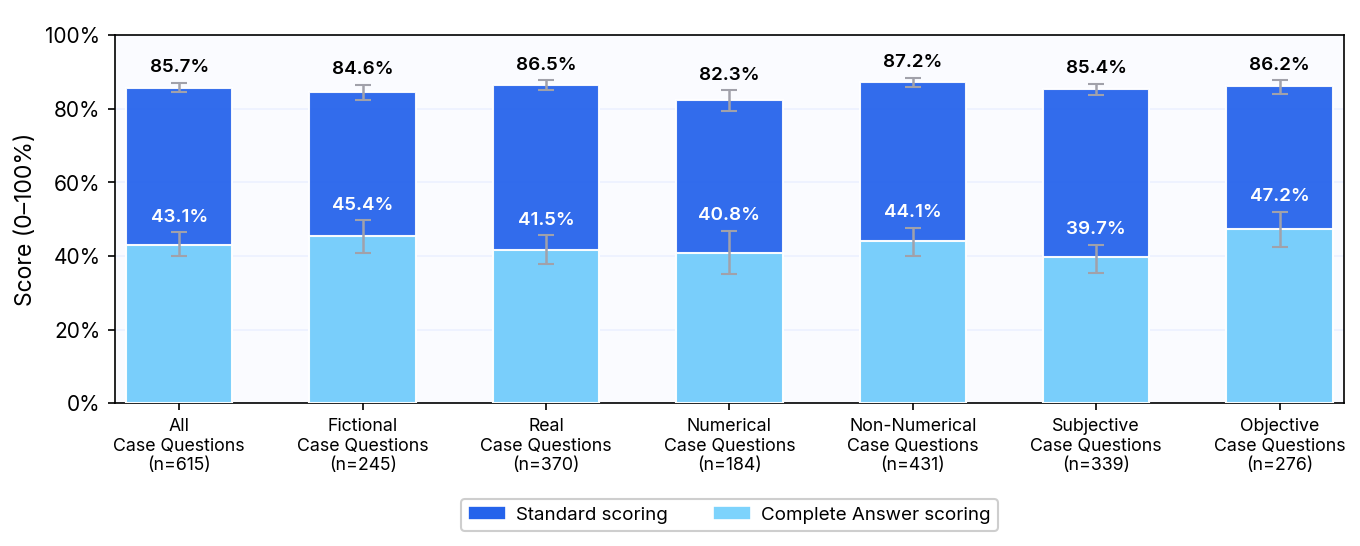}
\caption{Performance by case type and question type under Standard scoring and Complete Answer scoring. Grouped bars report mean scores with 95\% bootstrap confidence intervals for fictional vs.\ real cases, numerical vs.\ non-numerical questions, and subjective vs.\ objective questions, with each question averaged across GPT-5.4, Claude Sonnet~4.6, and Gemini~3 Flash Preview. Under Standard scoring all six strata fall in a narrow band; scores under Complete Answer scoring widen these gaps modestly but remain small relative to between-discipline variation documented in \figref{fig:complete_answer}.}
\label{fig:question_strata}
\end{Figure}

\refstepcounter{subsection}\label{supp:fictional-real-delta}
\subsection*{\thesubsection\quad Fictional-vs-real performance delta by discipline}

The figure plots, for each discipline with sufficient questions in both arms, the mean frontier score on fictional case questions minus the mean on real case questions, surfacing the Finance discipline as a major outlier as discussed in the main text.

\begin{Figure}
\centering
\includegraphics[width=\textwidth]{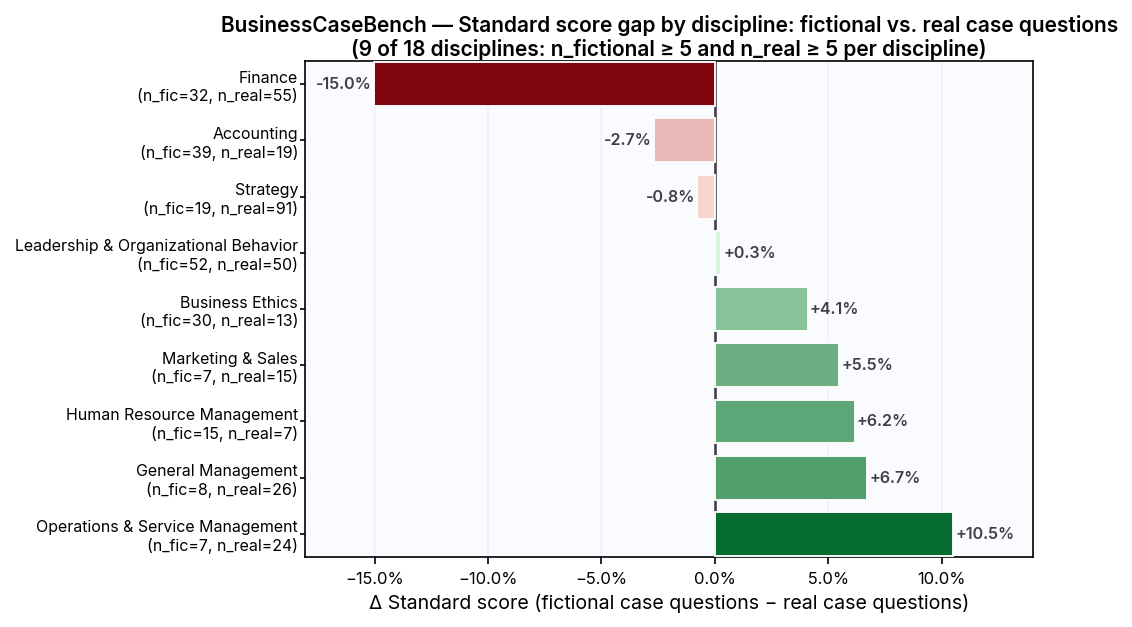}
\caption{Within-discipline gap between fictional and real case questions under Standard scoring. For each discipline with sufficient questions in both arms ($n_{\mathrm{fictional}} \geq 5$ and $n_{\mathrm{real}} \geq 5$), bars show the mean Standard score on fictional case questions minus the mean Standard score on real case questions, averaged across GPT-5.4, Claude Sonnet~4.6, and Gemini~3 Flash Preview. Positive values indicate higher scores on fiction; Finance is a pronounced negative outlier (fiction scores far below real), whereas most other disciplines show small fictional case advantages or near parity.}
\label{fig:fiction_vs_real_delta}
\end{Figure}

\refstepcounter{subsection}\label{supp:metadata-regression}
\subsection*{\thesubsection\quad Score variation by question type, discipline, and case}

To complement the descriptive strata and discipline comparisons in the main text, we fit descriptive ordinary least squares models predicting each question’s mean frontier Standard score (average of GPT-5.4, Claude Sonnet~4.6, and Gemini~3 Flash Preview). Scores lie on the 0–1 scale, so coefficients can be read as percentage-point differences in average score. The independent variables are binary indicators for numerical, subjective, and fictional-case questions, together with discipline fixed effects coded using sum-to-zero contrasts. Under this coding, discipline coefficients represent deviations from the adjusted average discipline effect. Disciplines represented by fewer than five questions (Social Enterprise, $n=1$; Management Communications, $n=4$) are pooled into a single Other category before fitting: cells this sparse produce high-leverage observations whose robust standard errors are unreliable, so the pooled Other term is included only as a control and its coefficient is not interpreted. Because multiple questions from the same case may share unobserved difficulty, inference uses standard errors clustered by case (238 clusters; $N = 615$).

\sitabref{tab:metadata-nested-r2} reports nested models. Taken at face value, question-type tags alone explain 3.1\% of question-level score variation, discipline alone explains 6.1\%, and both together explain 8.0\%. On an adjusted $R^{2}$ basis, the question-type tags explain 2.6\%, discipline 3.6\%, and the two together 5.1\%. Each adds information beyond the other. Adding discipline after the question-type tags raises adjusted $R^{2}$ by 0.025 (clustered Wald $F(16,237)=3.11$, $p < 0.001$), and adding the question-type tags after discipline raises it by 0.015 ($F(3,237)=4.16$, $p = 0.007$). Discipline keeps a modest edge in adjusted $R^{2}$ (3.6\% vs.\ 2.6\%), but both sets of labels leave most differences between harder and easier questions to case- and question-specific demands that coarse metadata do not capture. As a stronger check, we asked how well each set of labels predicts scores it was not fitted on, using cross-validation ($10$-fold, $50$ repeats). When entire cases are held out, so that the model must predict scores for cases it has never seen, the question-type tags predict slightly better than discipline (mean out-of-sample $R^{2}$ of about $0.010$ vs.\ $-0.004$, where the negative value means the predictions are no better than guessing the overall average score). Discipline is essentially a property of the case, spread across many small categories, so what it explains within this sample does not carry over to new cases. We therefore read discipline and question type as each carrying a detectable but small association with difficulty, rather than as evidence that discipline is the stronger predictor in general.


\begin{table}[H]
\caption{Nested ordinary least squares models of the mean frontier Standard score
(average of GPT-5.4, Claude Sonnet~4.6, and Gemini~3 Flash Preview from the primary
experiments; $0$--$1$ scale) on question-type strata and discipline ($N = 615$ questions).
Discipline uses effect (sum-to-zero) coding, with disciplines of fewer than five questions
pooled into a single Other category. Adjusted $R^{2}$ is reported alongside raw $R^{2}$
because the discipline block carries many more parameters than the three question-type
indicators. $\Delta$Adj.\ $R^{2}$ is the gain in adjusted $R^{2}$ from adding a block of
predictors, and $p$ is from a joint Wald $F$ test with standard errors clustered by case
(238 clusters).}\label{tab:metadata-nested-r2}%
\begin{tabular*}{\textwidth}{@{\extracolsep\fill}lccccc@{\extracolsep\fill}}
\hline
Model / comparison & $R^{2}$ & Adj.\ $R^{2}$ & $\Delta$Adj.\ $R^{2}$ & $F$ (df) & $p$ \\
\hline
Question-type tags only & 0.031 & 0.026 & --- & --- & --- \\
Discipline only & 0.061 & 0.036 & --- & --- & --- \\
Question type + discipline & 0.080 & 0.051 & --- & --- & --- \\
\hline
Discipline beyond question type & --- & --- & 0.025 & 3.11 (16, 237) & $<0.001$ \\
Question type beyond discipline & --- & --- & 0.015 & 4.16 (3, 237) & 0.007 \\
\hline
\end{tabular*}
\end{table}

\sitabref{tab:metadata-regression-coefs} reports coefficients from the joint model. In that specification, numerical questions score lower than non-numerical ones, while subjective vs.\ objective and fictional vs.\ real framing are not distinguishable from zero once the other predictors are included. Relative to the overall discipline mean, Business \& Government Relations and Strategy score higher and Operations \& Service Management scores lower at $p < 0.05$; Decision Analysis is marginally higher ($p = 0.051$), whereas Finance is not distinguishable from zero ($p = 0.117$). These descriptive decompositions show that coarse metadata leave most question-level score variation unexplained. The case-level analysis below asks how much of that residual is shared within cases.


\begin{table}[H]
\small
\caption{Coefficients from the joint model of \sitabref{tab:metadata-nested-r2}, with the same
outcome, sample, and case-clustered standard errors. Effect coding compares each discipline to
the overall mean across disciplines. The pooled Other category comprises the disciplines with
fewer than five questions (Social Enterprise, $n=1$; Management Communications, $n=4$), which
avoids unstable estimates from near-empty cells. The Other coefficient is included as a control
and is not interpreted. Coefficients are in score points on the $0$--$1$ scale
(e.g., $-0.05$\,$\approx$\,5 percentage points).}\label{tab:metadata-regression-coefs}%
\begin{tabular*}{\textwidth}{@{\extracolsep\fill}lrrrr@{\extracolsep\fill}}
\hline
Term & Coef & SE & $t$ & $p$ \\
\hline
Intercept & 0.904 & 0.013 & 70.85 & $<0.001$ \\
Numerical (vs non-numerical) & $-0.059$ & 0.019 & $-3.17$ & 0.002 \\
Subjective (vs objective) & $-0.032$ & 0.017 & $-1.90$ & 0.058 \\
Fictional case (vs real) & $-0.010$ & 0.019 & $-0.52$ & 0.606 \\
\hline
\multicolumn{5}{l}{\textit{Discipline (vs overall mean)}} \\
Business \& Government Relations ($n=12$) & 0.066 & 0.018 & 3.76 & $<0.001$ \\
Operations \& Service Management ($n=31$) & $-0.066$ & 0.030 & $-2.21$ & 0.028 \\
Strategy ($n=110$) & 0.026 & 0.013 & 1.98 & 0.048 \\
Decision Analysis ($n=21$) & 0.081 & 0.041 & 1.96 & 0.051 \\
Marketing \& Sales ($n=22$) & $-0.068$ & 0.036 & $-1.89$ & 0.060 \\
Finance ($n=87$) & $-0.042$ & 0.027 & $-1.58$ & 0.117 \\
Information Technology ($n=12$) & 0.024 & 0.019 & 1.25 & 0.214 \\
Leadership \& Organizational Behavior ($n=102$) & $-0.022$ & 0.019 & $-1.18$ & 0.241 \\
Negotiation ($n=8$) & $-0.045$ & 0.051 & $-0.89$ & 0.374 \\
General Management ($n=34$) & $-0.024$ & 0.028 & $-0.87$ & 0.385 \\
Economics ($n=24$) & 0.024 & 0.032 & 0.74 & 0.458 \\
Entrepreneurship \& Innovation ($n=17$) & 0.024 & 0.036 & 0.68 & 0.499 \\
International Business ($n=7$) & 0.032 & 0.048 & 0.68 & 0.499 \\
Business Ethics ($n=43$) & 0.012 & 0.022 & 0.56 & 0.574 \\
Accounting ($n=58$) & $-0.015$ & 0.029 & $-0.54$ & 0.592 \\
Human Resource Management ($n=22$) & $-0.018$ & 0.035 & $-0.53$ & 0.599 \\
Other ($n=5$; pooled) & 0.011 & 0.025 & 0.45 & 0.652 \\
\hline
\end{tabular*}
\end{table}

To test whether the difficulty left unexplained by coarse metadata is tied to the underlying case, we compare how much scores vary between the $238$ source cases versus among questions within the same case. Questions drawn from the same case have moderately correlated difficulty (intraclass correlation $0.22$), and average scores differ across cases by more than chance alone would produce ($F(237,377)=1.72$, $p < 0.001$). Knowing which case a question comes from also predicts its score far better than the coarse metadata labels do (adjusted $R^{2}=0.22$ vs.\ $0.05$ for question-type tags and discipline combined, \sitabref{tab:case-difficulty}). Accounting for case identity on top of the metadata model raises adjusted $R^{2}$ from $0.05$ to $0.24$, and the case effects remain jointly significant after the metadata block ($F(236,359)=1.64$, $p < 0.001$), so the case-level signal is not simply the metadata labels restated. Even so, nearly half of the variation in scores occurs among questions within the same case, and the question-type tags explain only about $1\%$ of that remaining variation. Difficulty is therefore partly a property of the case, which motivates the case-clustered standard errors above, but it remains substantially a property of the individual question.


\begin{table}[H]
\caption{How much question-level score variation is explained by case versus by coarse metadata.
The outcome, sample, and metadata model (question type plus pooled discipline) are as in
\sitabref{tab:metadata-nested-r2}. The case intraclass correlation (ICC) is from a one-way ANOVA
with unequal cluster sizes across the $238$ cases (60 contribute a single question and the rest
contribute between two and ten). Case fixed effects use one indicator per case. The last row
reports the adjusted $R^{2}$ gain from adding case identity to the metadata model, and the
nested $F$ tests whether that gain is jointly zero. Its numerator carries 236 rather than 237
degrees of freedom because fictional vs.\ real status is constant within a case and is absorbed
by the case indicators. A sensitivity analysis restricting to cases with at least $k$ questions
is reported in the text below.}\label{tab:case-difficulty}%
\begin{tabular*}{\textwidth}{@{\extracolsep\fill}lcccccc@{\extracolsep\fill}}
\hline
Model / comparison & ICC & $R^{2}$ & Adj.\ $R^{2}$ & $\Delta$Adj.\ $R^{2}$ & $F$ (df) & $p$ \\
\hline
Case (ANOVA) & 0.219 & --- & --- & --- & --- & --- \\
\hline
Metadata (question type + discipline) & --- & 0.080 & 0.051 & --- & --- & --- \\
Case fixed effects & --- & 0.520 & 0.218 & --- & 1.72 (237, 377) & $<0.001$ \\
Metadata + case fixed effects & --- & 0.558 & 0.243 & --- & --- & --- \\
\hline
Case beyond metadata & --- & --- & --- & 0.192 & 1.64 (236, 359) & $<0.001$ \\
\hline
\end{tabular*}
\end{table}

Because cases contribute unequal numbers of questions (60 of 238 cases contribute only one question, while the remainder contribute between two and ten), we recompute the case ICC and case fixed-effects fit after successively dropping cases below each size threshold. Restricting to the 178 cases with at least two questions yields ICC $=0.211$ and adjusted $R^{2}=0.211$ ($N=555$). Raising the threshold to three, four, or five questions per case leaves the ICC in a narrow 0.20--0.23 band and the adjusted $R^{2}$ in a 0.20--0.22 band. The case-level association is therefore not an artifact of singleton cases or of the unequal cluster-size distribution.

\refstepcounter{subsection}\label{supp:cross-model-complementarity}
\subsection*{\thesubsection\quad Cross-model complementarity}

The table summarizes how much performance rises when one takes, per question, the best score among the frontier trio (oracle ceiling) and how often all three models jointly fail the same questions or rubric criteria to gauge ``universal hardness''.


\begin{table}[H]
\caption{Cross-model oracle ceiling and universal hardness incidence for the frontier
trio (GPT-5.4, Claude Sonnet~4.6, and Gemini~3 Flash Preview). Universal hardness
threshold is Standard scoring $\leq 0.70$ on a question by all three frontier models.
}\label{tab:oracle-hardness}%
\begin{tabular*}{\columnwidth}{@{\extracolsep\fill}lcc@{\extracolsep\fill}}
\hline
Metric & Count & Share \\
\hline
\multicolumn{3}{l}{\textit{Question-level analysis ($N = 615$)}} \\
Questions scored by all three models & 615 & 100\% \\
Universally hard (max Standard scoring across all frontier models $\leq 70\%$) & 43 & 7.0\% \\
\hline
\multicolumn{3}{l}{\textit{Criterion-level analysis}} \\
Total rubric criterion instances & 8{,}369 & 100\% \\
Never-full-credit criteria (all three frontier models score 0) & 521 & 6.2\% \\
\hline
Metric & Count & Standard Score \\
\hline
Best single model (Claude Sonnet~4.6) & --- & 88.4\% \\
Oracle ceiling (per-question max, averaged) & --- & 92.8\% \\
Oracle lift over best single model & --- & \textbf{$+$4.5 pp} \\
\hline
\end{tabular*}
\end{table}

\refstepcounter{subsection}\label{supp:extended-onet-taxonomy-results}
\subsection*{\thesubsection\quad Extended O*NET taxonomy results}

These supplementary analyses extend the main-text IWA results (\figref{fig:iwa_heatmap}) to both O*NET occupations and the finer Detailed Work Activity (DWA) taxonomy.

\subsubsection*{Implied AI impact on O*NET occupations}

The occupation chart (\sifigref{fig:ai_impact_on_onet_occupations}) groups O*NET jobs by how densely the benchmark covers their relevant IWAs and reports mean frontier scores within each coverage band, demonstrating the implied AI impact on O*NET occupations according to the benchmark.

\begin{Figure}
\centering
\includegraphics[width=0.72\textwidth]{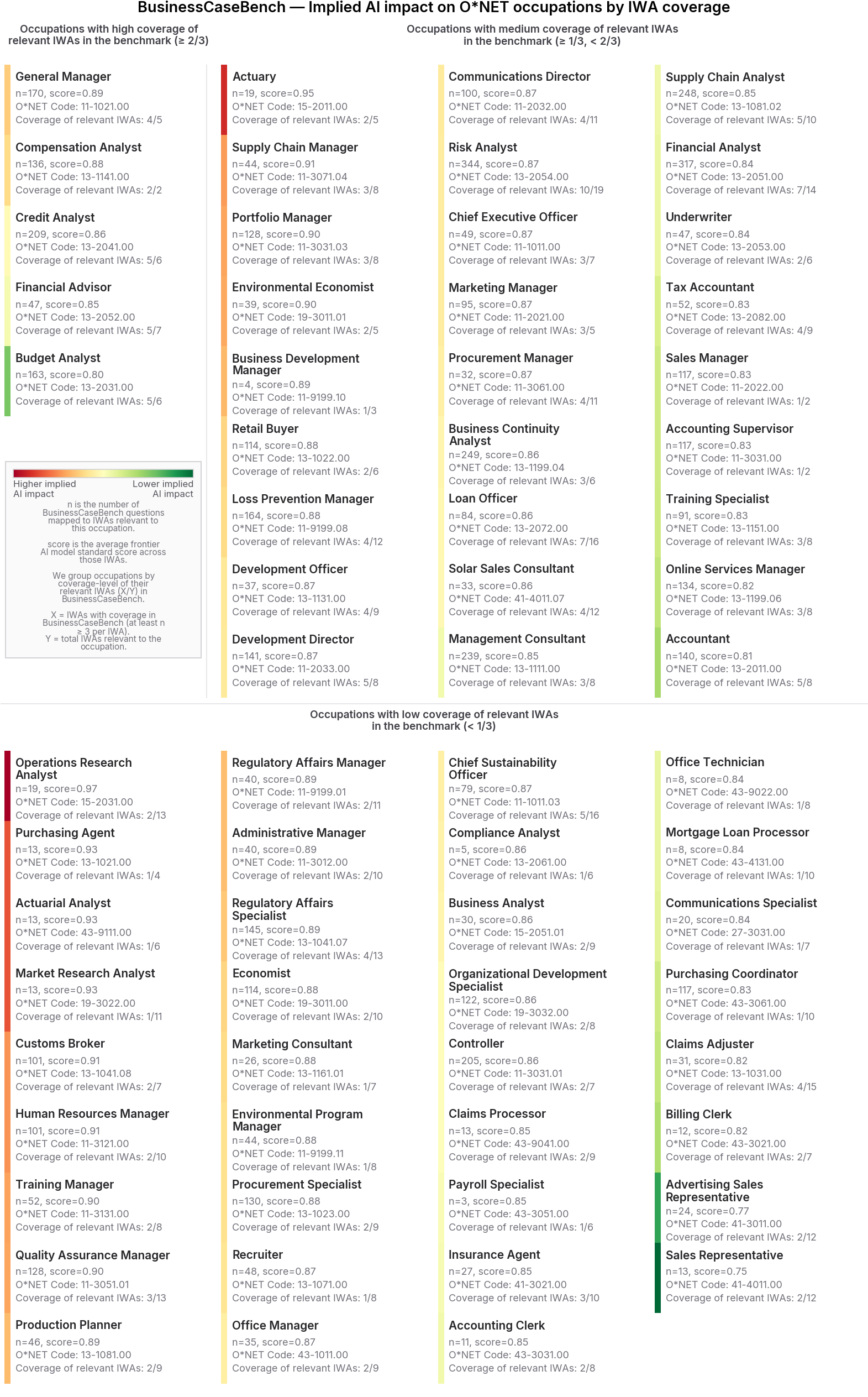}
\caption{Frontier AI performance on O*NET occupations represented in the benchmark, grouped by breadth of relevant Intermediate Work Activity (IWA) coverage. Occupations are partitioned into high, medium, and low coverage bands according to how many IWAs relevant to the occupation are represented sufficiently in the benchmark ($n \geq 3$); within each band, color bars encode the mean score under Standard scoring averaged across GPT-5.4, Claude Sonnet~4.6, and Gemini~3 Flash Preview on benchmark questions linked to relevant IWAs for that occupation, with the total number of contributing questions annotated. Occupations are sorted by implied AI impact within each coverage band. The figure summarizes which occupations the benchmark covers most densely and where implied AI impact concentrates at the job level.}
\label{fig:ai_impact_on_onet_occupations}
\end{Figure}

\subsubsection*{Implied AI impact by detailed work activity}

The DWA heatmap (\sifigref{fig:ai_impact_by_dwa}) extends the main-text IWA view to the finest O*NET work-activity level.

\begin{Figure}
\centering
\includegraphics[width=0.9\textwidth]{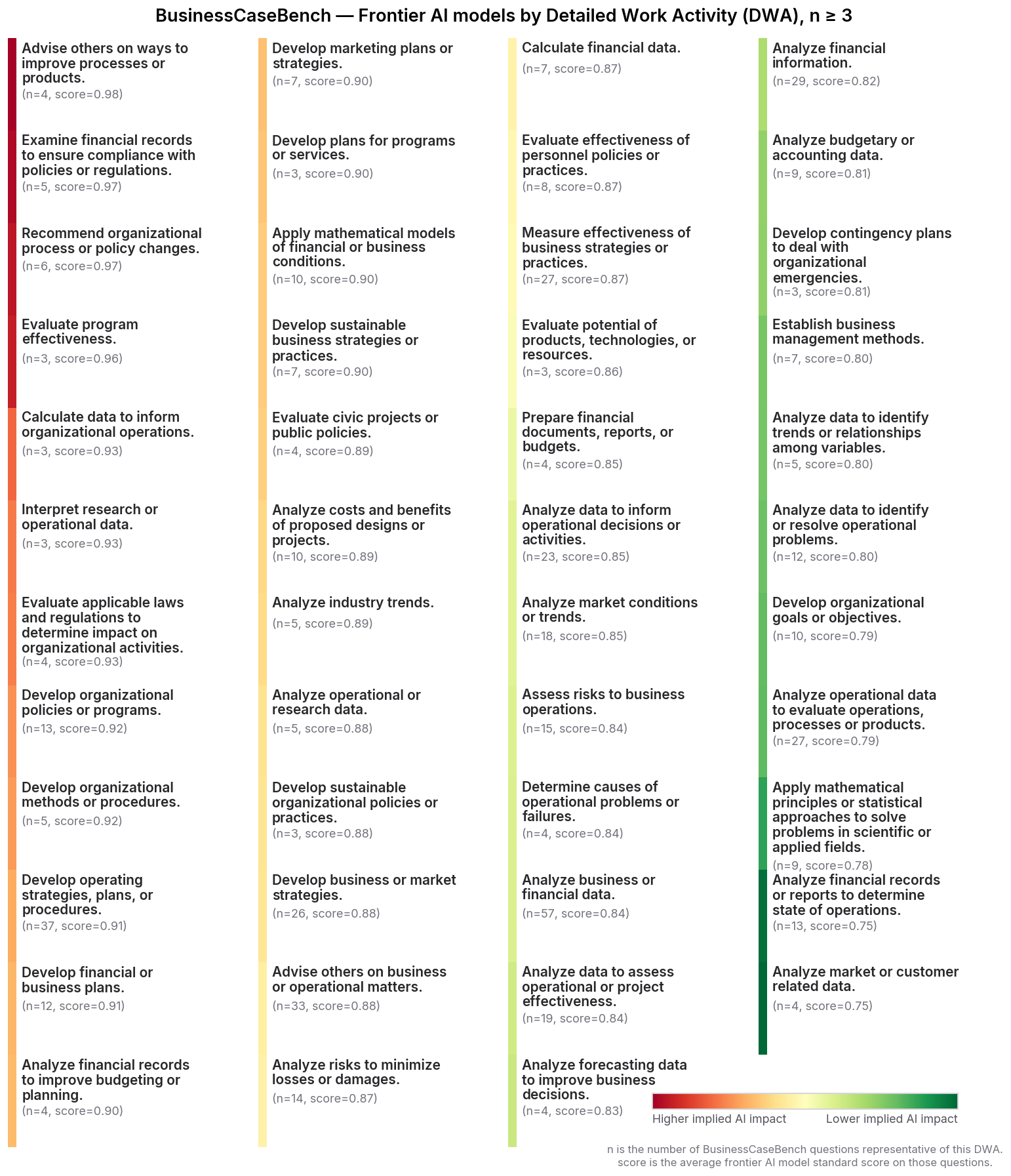}
\caption{Frontier AI performance by O*NET Detailed Work Activities (DWAs). Each cell is a DWA with at least three benchmark questions ($n \geq 3$); color encodes the mean score under Standard scoring averaged across GPT-5.4, Claude Sonnet~4.6, and Gemini~3 Flash Preview on those questions, with $n$ and score annotated. DWAs are sorted by implied AI impact. The finer DWA taxonomy reveals additional granularity in which specific professional subtasks resist full rubric satisfaction relative to aggregate IWA-level patterns.}
\label{fig:ai_impact_by_dwa}
\end{Figure}
\refstepcounter{section}\label{supp:validity-limitations-reproducibility}
\section*{\thesection\quad Validity, limitations, and reproducibility}
\setcounter{subsection}{0}

\refstepcounter{subsection}\label{supp:contamination-evidence}
\subsection*{\thesubsection\quad Contamination evidence}

We audit verbatim exposure of evaluation materials using InfiniGram \citep{Liu2024InfiniGram}, which supports efficient substring search over large, commonly used web-crawled pre-training corpora (C4 \citep{raffel2020exploring}, The Pile \citep{pile}, RedPajama \citep{together2023redpajama}, Dolma \citep{soldaini-etal-2024-dolma}, and DCLM \citep{li2025datacomplmsearchgenerationtraining}). For each case and its instructor case solution, we draw three random substrings of 5--10 tokens and query whether each appears in those indexes. We find no hits for the cases used in the evaluation pipeline, so the indexed corpora provide no evidence of verbatim contamination on this pre-training contamination probe.

\refstepcounter{subsection}\label{supp:scope-of-coverage}
\subsection*{\thesubsection\quad Scope of coverage}

The benchmark is single-turn and each instance supplies full case context in one prompt with no clarification or iterative evidence gathering (contrast with interactive diagnostic or agentic work benchmarks). This is intentional by design as this is how cases are generally designed to be used and helps isolate improvement on knowledge work and analytical reasoning without confounding on tool use capabilities or conversational dialogue capabilities. It is English-only and grounded in the business school case-method tradition (long narrative cases with instructor-derived reference standards) so it measures analytic and advisory synthesis under that particular pedagogical format. Results should be read as capability under this controlled protocol for evaluating business knowledge and reasoning.

\refstepcounter{subsection}\label{supp:released-withheld-artifacts}
\subsection*{\thesubsection\quad Released vs. withheld artifacts and access pathway}

\paragraph{Released.} Evaluation harness code, pinned model identifiers, hyperparameters, and prompts (including those in \sisecref{supp:model-evaluation-protocol}), structure of benchmark metadata and rubrics, and aggregated model outputs and scores.

\paragraph{Withheld.} Full case and instructor case solution PDFs and text, which publisher and clearinghouse licenses prohibit redistributing openly.

\paragraph{Access pathway.} Qualified researchers can reconstruct the evaluation set by obtaining cases through standard business school case clearinghouses and instructor-access channels, then repeating the procedure on locally held materials. License restrictions reduce open reusability but also limit verbatim inclusion of proprietary cases in pre-training corpora---the same constraint that allows such cases to be a valid measure of progress on knowledge work with low contamination risk.

\end{document}